\documentclass{article}

\usepackage{PRIMEarxiv}

\usepackage[utf8]{inputenc} 
\usepackage[T1]{fontenc}    
\usepackage{hyperref}       
\hypersetup{
	colorlinks=true,
	linkcolor=blue,
	citecolor=blue,
	filecolor=magenta,      
	urlcolor=cyan,
}
\usepackage{url}            
\usepackage{booktabs}       
\usepackage{amsfonts}       
\usepackage{nicefrac}       
\usepackage{microtype}      
\usepackage{lipsum}
\usepackage{fancyhdr}       
\usepackage{graphicx}       
\usepackage{tabularray}
\usepackage{amsmath}        
\graphicspath{{media/}}     
\usepackage{subfigure}       
\usepackage{siunitx}
\usepackage[sort&compress]{natbib}

\usepackage{xcolor}
\usepackage{mathtools}
\usepackage{enumitem}
\usepackage[most]{tcolorbox}
\usepackage{wrapfig}

\usepackage{pifont}

\numberwithin{equation}{section}

\title{On the training of physics-informed neural operators for solving parametric partial differential equations
}

\author{
  Nanxi Chen$^{1}$,\;
  Chuanjie Cui$^{2}$,\;
  Airong Chen$^{1}$,\;
	Sifan Wang$^{3,\ast}$,\;
	Rujin Ma$^{1,\ast}$\\[8pt]
  $^{1}$College of Civil Engineering, Tongji University, Shanghai 200092\\
  $^{2}$Department of Engineering Science, University of Oxford, Oxford OX1 3PJ\\
	$^{3}$Institute for Foundations of Data Science, Yale University, New Haven, CT 06520
}

\begin{document}
\maketitle
\begingroup
\renewcommand{\thefootnote}{\fnsymbol{footnote}}
\footnotetext[1]{Correspondence to: \texttt{sifan.wang@yale.edu}; \texttt{rjma@tongji.edu.cn}.}
\endgroup
\setcounter{footnote}{0}

\begin{abstract}
Physics-informed neural operators (PINOs) aim to learn solution operators for partial differential equations by using the governing physics as supervision, rather than relying solely on paired input-output simulation data. By incorporating physical constraints into the training objective, PINOs combine the cross-instance generalization of neural operators with the data efficiency of physics-informed learning. Despite this promise, how to train PINOs efficiently and robustly remains less well-understood than the training of either data-driven neural operators or physics-informed neural networks (PINNs).
To bridge this gap, we examine key components of the PINO training pipeline, including architecture design, optimizer choice, loss balancing, and collocation-point sampling strategy. We study three representative operator backbones, Deep Operator Network (DeepONet), Fourier Neural Operator (FNO), and Continuous Vision Transformer (CViT), across five diverse parametric PDE systems. Our results show that CViT provides consistently strong and stable performance across the considered benchmarks.
Beyond architecture, we find that several optimization pathologies previously identified in PINN training naturally arise in PINOs, including gradient conflicts and causal violation. We also find that mitigation algorithms developed for PINNs remain effective in the PINO setting. We further compare physics-informed and data-driven training under different data regimes, revealing that a carefully designed physics-informed training pipeline can match, and in some cases, outperform purely data-driven neural operators.
Taken together, these findings provide a systematic empirical understanding of the optimization challenges in PINO training and inform a practical pipeline for efficient and robust physics-informed operator learning. Code and data are publicly available at \url{https://github.com/NanxiiChen/PI-CViT}.
\end{abstract}

\keywords{Neural operators \and Physics-informed machine learning \and  Partial differential equations}

\section{Introduction}
\label{sec:introduction}

Scientific computing relies heavily on numerical solvers for partial differential equations (PDEs) to model complex physical systems and predict their behavior under varying conditions. Classical methods based on finite elements, finite volumes, and spectral methods have matured over decades and can provide highly accurate solutions for individual PDE instances. However, they become prohibitively expensive in high-throughput settings, where the same PDE family must be solved repeatedly with different parameters, initial conditions, or boundary conditions. The need for such repeated solutions arises across many scientific and engineering applications, including design optimization~\cite{lassilaParametricFreeformShape2010,wangmultiobjectaerodynamicOptimization2021}, materials discovery and screening~\cite{curtarolo2013high,jainMaterialsProject2013}, multiscale modeling~\cite{chatzopoulosPhysicsAwareNeuralImplicit2024,shyy2011surrogate,white2019multiscale}, reliability analysis~\cite{meng2023pinnform,li2005dynamic}, and real-time prediction and control~\cite{tomasetto2026real,prud2002mathematical,hwang2022solving}.

Scientific machine learning addresses this challenge by learning surrogate models that approximate expensive PDE solutions with efficient inference~\cite{brunton2024promising,karniadakisPhysicsinformedMachineLearning2021}, among which neural operators have emerged as a particularly powerful class~\cite{azizzadenesheli2024neural,kovachkiNeuralOperatorLearning2023}. Rather than solving each PDE instance independently, neural operators learn mappings from parametric inputs to solution fields and generalize across varying parameter configurations and discretizations. Representative architectures include the Deep Operator Network (DeepONet)~\cite{luLearningNonlinearOperators2021}, the Fourier Neural Operator (FNO)~\cite{liFourierNeuralOperator2021}, the Continuous Vision Transformer (CViT)~\cite{wang2024cvit}, among others~\cite{hao2023gnot,li2022transformer,cao2024laplace,rahman2022uno,li2023geometry,wu2025tante,seidman2022nomad,wen2022ufno}. These models have shown strong performance across a broad spectrum of applications, including surrogate modeling for fluid dynamics~\cite{li2022fourier,wang2024prediction}, microstructural modeling~\cite{jin2025characterization,bonnevilleAcceleratingPhaseField2025,oommenLearningTwophaseMicrostructure2022},  material deformation~\cite{koric2024deep}, and climate modeling~\cite{pathak2022fourcastnet}.

At the same time, physics-informed neural networks (PINNs) offer a complementary route to solving PDEs by incorporating the governing equations directly into the training loss~\cite{raissiPhysicsinformedNeuralNetworks2019a,karniadakisPhysicsinformedMachineLearning2021}. By minimizing PDE residuals together with initial and boundary condition terms at freely sampled collocation points, PINNs can be trained without labeled solution data, which is especially attractive when high-fidelity simulations are scarce or expensive. Although physics-informed learning is typically more challenging to optimize than supervised learning, the prospect of data-free training has motivated extensive work on new architectures~\cite{wangUnderstandingMitigatingGradient2021,wangPirateNetsPhysicsinformedDeep2024,wang2025kolmogorov}, sampling strategies~\cite{wuComprehensiveStudyNonadaptive2023,daw2022rethinking}, weighting schemes~\cite{wangUnderstandingMitigatingGradient2021,wangRespectingCausalityTraining2024}, optimizer design~\cite{wangGradientAlignmentPhysicsinformed2025a}, and other training techniques~\cite{chen2025sharp,chen2025enforcing,chiu2026scalepinn} for improving robustness and efficiency, as well as scientific applications spanning fluid mechanics~\cite{caiPhysicsinformedNeuralNetworks2021b}, phase-field materials modeling~\cite{chenPFPINNsPhysicsinformedNeural2025,chen2025sharp}, solid mechanics~\cite{wu2024pinnenhancedmultiscale,hu2024pinnsolid}, and geophysics~\cite{wang2025deeplearningflowlaw,rasht2022pinnwater}.

These two paradigms, although largely developed in parallel, are naturally complementary. Neural operators generalize across parametric PDE families, yet their standard supervised formulation typically requires large volumes of high-fidelity simulation data~\cite{bonnevilleAcceleratingPhaseField2025}. Physics-informed learning removes the need for labeled data, but in its conventional PINN formulation the model solves only a single PDE instance per training run. Combining operator architectures with physics-informed training produces physics-informed neural operators (PINOs)~\cite{wangLearningSolutionOperator2021,liPhysicsinformedNeuralOperator2021,wangLongtimeIntegrationParametric2023}, which offer the prospect of uniting the generalization capability of neural operators with the data-free property of physics-informed learning. Early explorations have demonstrated the feasibility of this approach on specific PDE systems, with applications to heat transfer~\cite{koricDatadrivenPhysicsinformedDeep2023a}, solid and structural mechanics~\cite{kaewnuratchadasorn2024physics,tian2026pinobending,eshaghi2025variational}, wave propagation~\cite{ma2026effectivepinowavefields}, and phase-field materials modeling~\cite{chen2026pfpino}.

However, as PINOs incorporate the physics-informed training objective, they are susceptible to the optimization pathologies that have been well documented in the PINN literature, including spectral bias~\cite{wangEigenvectorBiasFourier2021}, gradient imbalance among competing loss terms~\cite{wangUnderstandingMitigatingGradient2021}, causal violation in time-dependent problems~\cite{wangRespectingCausalityTraining2024}, and gradient conflicts~\cite{wangGradientAlignmentPhysicsinformed2025a}. These issues may even be exacerbated by the increased complexity of learning mappings between function spaces. Furthermore, the literature still lacks a unified and practically grounded understanding of how to train PINOs reliably across different PDE families and operator backbones. This naturally leads to two paired questions when extending from PINNs to PINOs:
\begin{center}

\itshape

To what extent do PINN training strategies transfer effectively to PINOs, and how well do PINOs preserve the data-free advantages of physics-informed training?

\end{center}

In this work, we investigate these questions through a systematic empirical study of physics-informed neural operator training. Our main contributions are summarized as follows:
\begin{itemize}[leftmargin=*]
\item We evaluate PINO across five parametric PDE systems and three representative operator backbones (PI-CViT, PI-DeepONet, and PI-FNO), and find that PI-CViT consistently delivers the strongest performance.
\item We find that key optimization pathologies previously identified in PINN training, such as gradient imbalance, causal violation, and gradient conflicts, naturally arise in PINO training.
\item We demonstrate that training strategies developed for PINNs, including GradNorm weighting, causal weighting, and the SOAP optimizer, effectively improve the predictive accuracy of PINOs.
\item We show that a well-resolved purely physics-informed training pipeline can match or surpass purely data-driven training across different labeled-data scales.
\item We distill these findings into a structured training pipeline and practical guidelines for physics-informed neural operators.
\end{itemize}
The remainder of the paper is organized as follows. Section~\ref{sec:operator_learning} formulates the conventional and physics-informed operator-learning objectives, and Section~\ref{sec:method} then develops our training framework for PINOs, including architecture design, weighting and optimization strategies, and practical considerations. Section~\ref{sec:results} evaluates this framework across a suite of benchmarks while comparing different operator backbones, followed by comprehensive ablation studies that isolate the contribution of each key component. Finally, Section~\ref{sec:conclusion} summarizes the main findings and outlines directions for future work.

\section{Physics-informed operator learning}
\label{sec:operator_learning}
Physics-informed operator learning aims to approximate the solution map of a class of PDEs while explicitly incorporating the governing physical laws into the learning process. To formalize this setting, we consider a family of PDEs parameterized by an input function $u \in \mathcal{U}$. For each instance $u \in \mathcal{U}$, the corresponding solution field $s(\mathbf{x}, t; u)$ is defined on a spatial domain $\Omega$ and, for time-dependent problems, on a temporal interval $[0,T]$. The governing PDE, initial condition, and boundary condition are written in a general form as
\begin{alignat}{2}
    \mathcal{N}(s(\mathbf{x}, t; u); u) &= 0, \quad 
    && (\mathbf{x}, t) \in \Omega \times [0,T], \\
    \mathcal{I}(s(\mathbf{x}, t; u); u) &= 0, \quad 
    && (\mathbf{x}, t) \in \Omega \times \{0\}, \\
    \mathcal{B}(s(\mathbf{x}, t; u); u) &= 0, \quad 
    && (\mathbf{x}, t) \in \partial\Omega \times [0,T],
\end{alignat}
where $\mathcal{N}(\cdot\,; u)$ denotes the PDE operator for the instance defined by $u$, and $\mathcal{I}(\cdot\,; u)$ and $\mathcal{B}(\cdot\,; u)$ denote the corresponding initial and boundary operators.
The solution operator $\mathcal{G}$ maps each input function $u$ to its corresponding solution field $s(\cdot\,; u)$, defining a mapping between the input function space $\mathcal{U}$ and the solution space $\mathcal{S}$:
\begin{equation}
	\mathcal{G}: \mathcal{U} \to \mathcal{S}, \qquad u \mapsto s(\cdot\,; u),
\end{equation}
and our goal is to learn a parametric approximation $\mathcal{G}_{\boldsymbol{\theta}}$ such that
\begin{equation}
	\mathcal{G}_{\boldsymbol{\theta}} : \mathcal{U} \to \mathcal{S}, \qquad \mathcal{G}_{\boldsymbol{\theta}}(u) \approx \mathcal{G}(u) = s(\cdot\,; u), \quad \forall u \in \mathcal{U}.
\end{equation}
In practice, this learning problem is formulated as an optimization over the model parameters:
\begin{equation}
	\boldsymbol{\theta}^{\ast} = \arg\min_{\boldsymbol{\theta}} \, \mathcal{L}(\mathcal{G}_{\boldsymbol{\theta}}),
\end{equation}
where $\mathcal{L}$ denotes a training loss that measures how well $\mathcal{G}_{\boldsymbol{\theta}}$ approximates the true solution operator.

The most straightforward way to train a neural operator is through supervised learning, in which the loss is evaluated on labeled trajectories that provide input-output pairs. Given a set of input functions $\{u_j\}_{j=1}^M$ and reference solutions sampled at coordinates $\{(\mathbf{x}_i, t_i)\}_{i=1}^{N_{\text{data}}}$, the standard supervised loss is defined as
\begin{equation}
	\mathcal{L}_{\text{data}}(\mathcal{G}_{\boldsymbol{\theta}}) = \frac{1}{MN_{\text{data}}} \sum_{j=1}^M \sum_{i=1}^{N_{\text{data}}} \left| \mathcal{G}_{\boldsymbol{\theta}}(u_j)(\mathbf{x}_i, t_i) - s_{\text{ref}}(\mathbf{x}_i, t_i; u_j) \right|^2,
\end{equation}
where $s_{\text{ref}}(\cdot\,; u_j)$ denotes the reference solution for the $j$-th input instance. In this conventional data-driven setting, the governing physics is enforced only implicitly through the labeled solution trajectories.

However, as discussed in the Introduction, generating high-fidelity solution trajectories can be expensive, and fitting data alone may not adequately enforce the governing physical constraints. Physics-informed training offers an alternative by directly minimizing PDE residuals at freely sampled collocation points~\cite{raissiPhysicsinformedNeuralNetworks2019a,karniadakisPhysicsinformedMachineLearning2021a}. Specifically, the PDE residual, together with the initial  and boundary condition losses, can be incorporated into the training objective to guide the model toward physically consistent solutions. Given PDE residual, initial condition, and boundary collocation points $\{(\mathbf{x}_i, t_i)\}_{i=1}^{N_r}$, $\{(\mathbf{x}_i, 0)\}_{i=1}^{N_{\text{ic}}}$, and $\{(\mathbf{x}_i, t_i)\}_{i=1}^{N_{\text{bc}}}$, sampled from $\Omega \times [0,T]$, $\Omega \times \{0\}$, and $\partial\Omega \times [0,T]$, respectively, the individual loss components are defined as
\begin{align}
    \mathcal{L}_r
    &=
    \frac{1}{M N_r}
    \sum_{j=1}^M \sum_{i=1}^{N_r}
    \left|
    \mathcal{N}(\mathcal{G}_{\boldsymbol{\theta}}(u_j)(\mathbf{x}_i,t_i);u_j)
    \right|^2,
    \\
    \mathcal{L}_{\text{ic}}
    &=
    \frac{1}{M N_{\text{ic}}}
    \sum_{j=1}^M \sum_{i=1}^{N_{\text{ic}}}
    \left|
    \mathcal{I}(\mathcal{G}_{\boldsymbol{\theta}}(u_j)(\mathbf{x}_i,0);u_j)
    \right|^2,
    \\
    \mathcal{L}_{\text{bc}}
    &=
    \frac{1}{M N_{\text{bc}}}
    \sum_{j=1}^M \sum_{i=1}^{N_{\text{bc}}}
    \left|
    \mathcal{B}(\mathcal{G}_{\boldsymbol{\theta}}(u_j)(\mathbf{x}_i,t_i);u_j)
    \right|^2.
\end{align}
A weighted combination of these terms yields the overall training objective:
\begin{equation}
	\mathcal{L} = w_r \mathcal{L}_r + w_{\text{ic}} \mathcal{L}_{\text{ic}} + w_{\text{bc}} \mathcal{L}_{\text{bc}} + w_{\text{data}} \mathcal{L}_{\text{data}},
	\label{eq:total_loss}
\end{equation}
The weight coefficients $w_r$, $w_{\text{ic}}$, $w_{\text{bc}}$, and $w_{\text{data}}$ balance the contributions of the different loss terms. The derivative terms in the PDE residual $\mathcal{L}_r$ can be computed via automatic differentiation (AD) or finite-difference (FD) approximations, depending on the model architecture and the collocation-point distribution. The supervised term $\mathcal{L}_{\text{data}}$ is optional and can be omitted when labeled trajectories are scarce or unavailable, yielding a purely physics-informed training objective.

However, the composite objective in Eq.~\eqref{eq:total_loss} can be challenging to optimize. For example, the PDE residual loss $\mathcal{L}_r$ may produce gradients with vastly different magnitudes from other components and varies dynamically during training; the gradient directions of different loss components may conflict; and for time-dependent problems, the model may violate causality by fitting later states before learning earlier time segments. In the next section, we develop a comprehensive training framework that addresses these challenges for a robust and efficient optimization of physics-informed neural operators

\section{Method}
\label{sec:method}

The physics-informed operator-learning problem introduced above places coupled demands on the model architecture and the training procedure. On the modeling side, the network must encode a parametric input function and represent the corresponding solution in a form that can be queried accurately and differentiably across the spatiotemporal domain. On the training side, the objective combines several heterogeneous terms, including PDE residual, initial-condition, boundary-condition, and optional supervised data losses, whose magnitudes and gradients can differ substantially. These challenges motivate a framework that pairs an expressive neural operator backbone with stabilization strategies for loss weighting, optimization, and collocation sampling.

\begin{figure}[t]
	\centering
	\includegraphics[width=\textwidth]{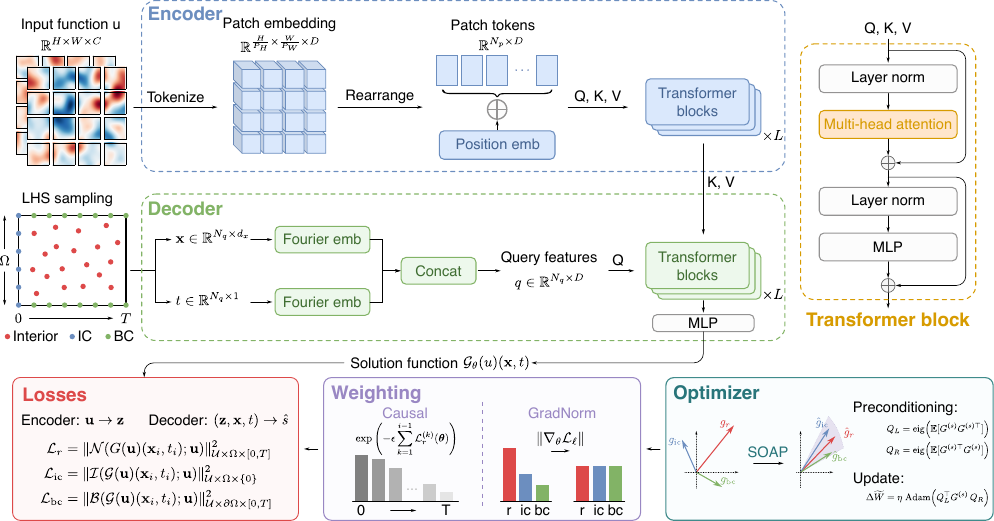}
	\caption{Overview of the training pipeline for physics-informed neural operators. The input function is encoded by a vision-transformer-based encoder into a latent representation. Given spatiotemporal query coordinates, the decoder attends to this latent representation and predicts the solution at the queried locations. The training objective combines multiple loss components, including PDE residual, initial-condition, and boundary-condition losses. The PDE residual loss is further modulated by a causal weighting scheme that emphasizes earlier time segments. A gradient-normalization weighting scheme balances the contributions of different loss terms, while the SOAP optimizer is used to mitigate gradient conflicts during training.}
	\label{fig:master_pipeline}
\end{figure}

In this section, we describe the main components of this framework. We first discuss the choice of neural operator architecture and motivate the use of a transformer-based coordinate decoder. We then introduce the loss-balancing and optimization strategies used to stabilize physics-informed training, followed by the sampling schemes and implementation details used in our experiments. An overview of the full training pipeline is shown in Figure~\ref{fig:master_pipeline}.

\subsection{Architecture design for physics-informed neural operators}
\label{sec:architecture}

\begin{figure}[t]
	\centering
	\includegraphics[width=\textwidth]{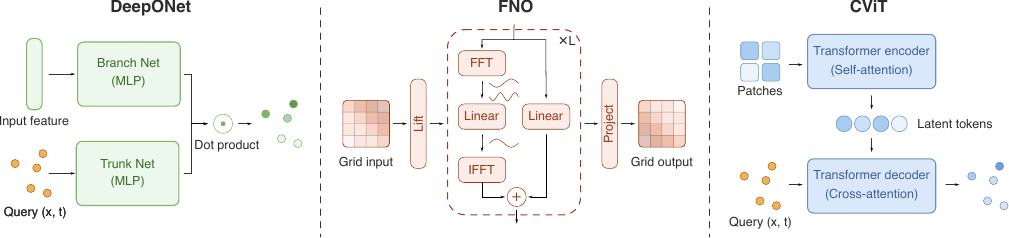}
	\caption{Schematic comparison of the three neural operator architectures studied in this work. DeepONet combines branch and trunk networks through an inner product for pointwise prediction. FNO applies Fourier layers on structured grids to capture global interactions in the spectral domain. CViT encodes the input field into latent tokens with a vision transformer and decodes the solution at arbitrary query coordinates through cross-attention.}
	\label{fig:arches}
\end{figure}

The choice of operator architecture is central to physics-informed training, as the model must translate a parametric input field into a solution representation that can be evaluated at arbitrary spatiotemporal query points. This requirement is particularly important when PDE residuals and constraint losses are computed at collocation points that may differ from the input discretization or from any available labeled data. With this motivation, we consider three representative neural operator backbones: DeepONet~\cite{luLearningNonlinearOperators2021}, FNO~\cite{liFourierNeuralOperator2021}, and CViT~\cite{wang2024cvit}. A schematic comparison of these architectures is provided in Figure~\ref{fig:arches}.

DeepONet and FNO represent two widely used approaches to operator learning. DeepONet uses separate branch and trunk networks to encode the input function and the query coordinates, respectively, and combines the resulting features through an inner product for pointwise prediction. This design naturally supports coordinate-based evaluation. FNO, by contrast, is built on Fourier layers and typically operates on structured grids, where global interactions are captured efficiently in the spectral domain. Both architectures have been widely studied and successfully applied to a range of operator-learning problems.

Despite these advantages, several limitations become relevant in the physics-informed setting considered here. DeepONet's fully connected branch--trunk design may be less effective in capturing complex spatiotemporal structure~\cite{zhu2023fourierdeeponet}, and its inner-product interaction can restrict the representation of intricate parametric dependencies~\cite{jin2022mionet}. FNO efficiently models global interactions on regular grids, but extending it to arbitrary query locations or irregular discretizations often requires additional embeddings or architectural modifications~\cite{liFourierNeuralOperator2023}. Moreover, its Fourier basis can be less convenient for strongly multiscale features or non-periodic problems~\cite{wen2022ufno,liPhysicsinformedNeuralOperator2021}.

These considerations motivate the use of an architecture that combines expressive input encoding with flexible coordinate-based decoding. CViT provides such a formulation by using a vision transformer to encode the parametric field into a latent representation and a cross-attention decoder to predict the solution at arbitrary spatiotemporal query coordinates. This structure is well suited to physics-informed operator learning, since residual, initial-condition, and boundary-condition losses can be evaluated at different sets of collocation points without requiring the output to be tied to a fixed grid. As discussed later in Section~\ref{sec:results}, CViT consistently delivers strong performance across the benchmark problems considered here relative to DeepONet and FNO. We therefore adopt CViT as the backbone of our physics-informed neural operator framework and refer to the resulting model as PI-CViT. Its architecture is detailed below.

\paragraph{Encoder.} Given an input field $u \in \mathbb{R}^{H\times W\times C}$, we first apply a convolutional patch-embedding layer to partition the input into non-overlapping patches and map them to patch tokens $\mathbf{u}_p \in \mathbb{R}^{N_p \times D_e}$. We then add fixed two-dimensional sinusoidal positional embeddings $\mathrm{PE} \in \mathbb{R}^{N_p \times D_e}$ to the patch tokens:
\begin{equation}
	\mathbf{u}_{\mathrm{pe}} = \mathbf{u}_p + \mathrm{PE}, \qquad \mathbf{u}_{\mathrm{pe}} \in \mathbb{R}^{N_p \times D_e}.
\end{equation}
where $P_H$ and $P_W$ are the patch height and width, $N_p = \frac{H}{P_H} \times \frac{W}{P_W}$ is the total number of patches, and $D_e$ is a hyperparameter controlling the encoder embedding dimension. 

The encoder then applies a stack of self-attention blocks,
\begin{alignat}{2}
	\hat{\mathbf{z}}_{l+1} &= \mathbf{z}_l + \mathrm{MSA}(\mathrm{LN}(\mathbf{z}_l)), \quad && \text{for } l=0,\dots,L_e-1,
	\\
	\mathbf{z}_{l+1} &= \hat{\mathbf{z}}_{l+1} + \mathrm{MLP}(\mathrm{LN}(\hat{\mathbf{z}}_{l+1})), \quad &&\text{for }  l=0,\dots,L_e-1,
\end{alignat}
where $\mathbf{z}_0 = \mathbf{u}_{\mathrm{pe}}$, $\mathrm{MSA}$ denotes a standard multi-head self-attention operation, and $\mathrm{MLP}$ denotes a feedforward network. The encoder output is a set of latent tokens $\mathbf{z}_{L_e} \in \mathbb{R}^{N_p \times D_e}$.

\paragraph{Decoder.} To evaluate the solution at a set of
query coordinates $\{(\mathbf{x}_i, t_i)\}_{i=1}^{N_q}$, the decoder constructs coordinate-dependent query features. Fourier features are applied separately to the spatial and temporal coordinates to capture their distinct roles in the PDE solution~\cite{wangEigenvectorBiasFourier2021,tancikFourierFeaturesLet2020}:
\begin{alignat}{2}
	\mathbf{q}_x(\mathbf{x}) &= [\sin(2\pi \mathbf{B}_x \mathbf{x}), \cos(2\pi \mathbf{B}_x \mathbf{x})], \quad && \mathbf{q}_x \in \mathbb{R}^{N_q\times (D_e/2)},
	\\
	\mathbf{q}_t(t) &= [\sin(2\pi \mathbf{B}_t t), \cos(2\pi \mathbf{B}_t t)], \quad &&\mathbf{q}_t \in \mathbb{R}^{N_q\times (D_e/2)},
\end{alignat}
where $\mathbf{B}_x$ and $\mathbf{B}_t$ are projection matrices that randomly sample frequencies from a Gaussian distribution, and $N_q$ is the number of query points.

The decoder then applies a stack of cross-attention blocks to attend over the latent encoder representation and produce solution representations at the queried coordinates:
\begin{alignat}{2}
	\hat{\mathbf{q}}_{l+1} &= \mathbf{q}_l + \mathrm{MCA}(\mathrm{LN}(\mathbf{q}_l), \mathrm{LN}(\hat{\mathbf{z}}_{L_e}), \mathrm{LN}(\hat{\mathbf{z}}_{L_e}))
	\quad &&\text{for } l=0,\dots,L_d-1,
	\\
	\mathbf{q}_{l+1} &= \hat{\mathbf{q}}_{l+1} + \mathrm{MLP}(\mathrm{LN}(\hat{\mathbf{q}}_{l+1})), \quad &&\text{for } l=0,\dots,L_d-1,
\end{alignat}
where $\mathbf{q}_0 = [\mathbf{q}_x(\mathbf{x}), \mathbf{q}_t(t)]$ is the initial query feature, $\hat{\mathbf{z}}_{L_e}\in\mathbb{R}^{N_p \times D_d}$ is the latent encoder output projected by a linear layer to match the decoder embedding dimension, and $\mathrm{MCA}$ denotes a standard multi-head cross-attention operation. Finally, an MLP head is applied to the output query tokens to obtain the predicted solution:
\begin{equation}
	\hat{s}(\mathbf{x},t;u) = \mathrm{MLP}(\mathrm{LN}(\mathbf{q}_{L_d})).
\end{equation}

\subsection{Loss formulation and weighting}
\label{sec:loss}

The physics-informed loss in Eq.~\eqref{eq:total_loss} combines several physical constraints, including PDE residuals, initial conditions, and boundary conditions. Although it is written as a single scalar objective, it behaves in practice as a multi-objective optimization problem. In the PINN literature, such composite objectives are known to suffer from gradient imbalance, where different loss terms generate gradients with vastly different magnitudes and the optimizer overemphasizes the dominant term~\cite{wangUnderstandingMitigatingGradient2021}. For time-dependent problems, another difficulty is causal violation: uniformly minimizing residuals over the entire time domain can encourage the model to fit later-time dynamics before earlier-time dynamics have been resolved~\cite{wangRespectingCausalityTraining2024}. These pathologies naturally carry over to PINOs, where the same physical constraints must be satisfied not for a single PDE instance, but across an entire family of input functions.

We therefore adopt two complementary weighting strategies inspired by prior PINN training methods. Gradient-norm-based weighting balances the contributions of different loss components, while causal weighting reweights the PDE residual across time so that earlier temporal segments are prioritized before later ones. Details of these strategies are described below.

\paragraph{Gradient-norm-based loss balancing.}

\begin{figure}[h]
	\centering
	\includegraphics{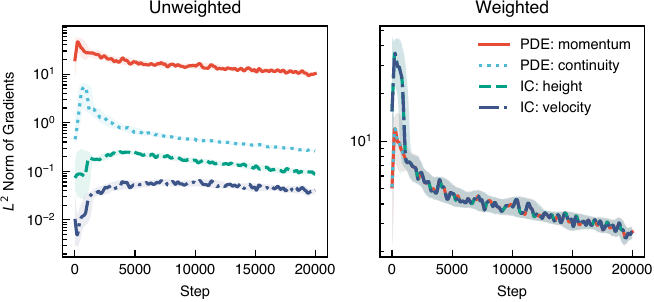}
	\caption{$L^2$ norms of the gradients of the four loss components (two PDE residual terms: momentum and continuity; two initial condition terms: height and velocity) during training on the shallow water benchmark with PI-DeepONet.
		\textbf{Left}: Unweighted losses. The momentum residual gradient dominates, exceeding the IC velocity gradient by more than two orders of magnitude throughout training.
		\textbf{Right}: GradNorm-weighted losses. The adaptive weighting scheme balances each gradient norm to ensure equitable contributions to the overall training dynamics.}
	\label{fig:gradnorms}
\end{figure}

The gradient imbalance discussed above is clearly visible in the shallow water benchmark. As shown in Figure~\ref{fig:gradnorms}, PI-DeepONet is trained with four coupled loss components: two PDE residual losses for the momentum and continuity equations, and two initial-condition losses for the height and velocity fields. Throughout training, the gradient norm of the momentum residual remains more than two orders of magnitude larger than that of the initial-velocity loss. Consequently, the optimization is biased toward reducing the dominant residual term, while neglecting to fit the initial conditions of the PDE system.

To mitigate this imbalance, we adopt a GradNorm-style weighting scheme~\cite{chenGradNormGradientNormalization2018,wangUnderstandingMitigatingGradient2021}. The idea is to adaptively rescale each loss component according to its gradient magnitude, so that no single term dominates the parameter update. Let $\mathcal{J}_{\mathcal{L}}$ denote the index set of all loss terms. At training step $s$, the raw weight for the $\ell$-th loss term is defined from its gradient as
\begin{align}
    g_\ell^{(s)}
        &= \nabla_{\boldsymbol{\theta}} \mathcal{L}_\ell^{(s)},
        \quad \ell \in \mathcal{J}_{\mathcal{L}},
        \\
    \hat{w}_\ell^{(s)}
        &= \frac{1}{\left\lVert g_\ell^{(s)}\right\rVert}
        \sum_{k\in \mathcal{J}_{\mathcal{L}}} \left\lVert g_k^{(s)} \right\rVert
        \label{eq:grad_norm_weighting}
\end{align}
where $g_\ell^{(s)}$ denotes the gradient of the $\ell$-th loss with respect to the model parameters $\boldsymbol{\theta}$ at step $s$. The raw weights $\hat{w}_\ell^{(s)}$ are then smoothed via an exponential moving average to reduce oscillations:
\begin{equation}
	w_\ell^{(s)} = \alpha_w \hat{w}_\ell^{(s)} + (1-\alpha_w) w_\ell^{(s-1)}, \quad w_\ell^{(0)} = 1,
\end{equation}
where $\alpha_w \in (0,1]$ controls the degree of smoothing. This scheme promotes balanced gradient contributions across all loss terms, leading to more stable and efficient training.

\paragraph{Causal weighting for time-dependent PDEs.}
\label{sec:causal}

Gradient-norm weighting balances different loss components, but it does not address the temporal structure inside the PDE residual itself. For time-dependent PDEs, the solution at time $t$ should be determined only by the initial/boundary conditions and the dynamics on earlier times, not by future states~\cite{wangRespectingCausalityTraining2024}. However, the standard residual loss averages collocation points over the whole time domain, treating early and late times equally. This can allow the optimizer to reduce later-time residuals before the earlier dynamics are accurately learned, leading to solutions that have small averaged residuals but violate the causal evolution of the PDE.

\begin{figure}[htbp]
	\centering
	\includegraphics{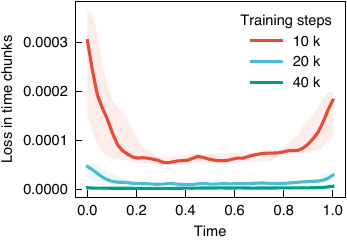}
	\caption{PDE residual loss per time segment at training steps 10k, 20k, and 40k for PI-CViT on the wave equation. Under causal weighting, the optimizer progressively satisfies earlier time segments before advancing to later ones, resulting in a uniform loss profile by 40k steps.}
	\label{fig:causal_loss_chunks}
\end{figure}

Following the causal weighting strategy developed for PINNs~\cite{wangRespectingCausalityTraining2024}, we impose a temporal ordering on the residual loss. Specifically, we partition the time interval $[0,T]$ into $N_t$ equal segments and assign each segment a weight determined by the cumulative residual error of all previous segments. Later segments are therefore emphasized only after the model has sufficiently reduced the residual at earlier times:
\begin{equation}
	w_{\text{causal}}^{(i)} = \exp\left(-\epsilon \sum_{k=1}^{i-1} \mathcal{L}_r^{(k)} (\boldsymbol{\theta}) \right),\quad \text{for } i=1,2,\ldots,N_t, \label{eq:causal_weighting}
\end{equation}
where $\epsilon$ is a hyperparameter controlling the strength of causality enforcement, and $\mathcal{L}_r^{(k)}$ is the MSE of the PDE residual at collocation points within the $k$-th time segment, averaged over the batch of input functions $\{u_j\}_{j=1}^M$. We increase $\epsilon$ when $0.99<w_{\text{causal}}^{(N_t)} < w_{\text{causal}, \text{max}}$ to progressively enforce causality as training proceeds. A representative example is shown in Figure~\ref{fig:causal_loss_chunks} for PI-CViT on the wave equation, where strongly time-dependent dynamics make causality-related training behavior particularly visible: at 10k steps the per-segment loss is still elevated at early times, but as training proceeds the causal weights guide the optimizer to resolve earlier segments first, and by 40k steps the loss is uniformly small across the entire temporal domain.

The ablation study in Section~\ref{sec:ablation_weighting} reveals that both the GradNorm and causal weighting schemes improve the predictive accuracy of PINOs, and their combination yields the best performance across the considered problems.

\subsection{Optimizer selection and gradient alignment}

\label{sec:optimizer}

Physics-informed neural operators are trained by minimizing a composite objective consisting of PDE residual, initial condition, boundary condition, and optional supervised data terms. Beyond the gradient magnitude imbalance discussed above, such a formulation also gives rise to directional conflicts among the loss components. When the gradients of competing objectives point in opposing directions, standard first-order optimizers such as Adam may struggle to find a descent direction that simultaneously reduces all loss terms---a phenomenon known as directional gradient conflict~\cite{wangGradientAlignmentPhysicsinformed2025a,liu2025config}.

We quantify this conflict using the gradient alignment score introduced in~\cite{wangGradientAlignmentPhysicsinformed2025a}. Let $\{g_\ell^{(s)}\}_{\ell \in \mathcal{J}_{\mathcal{L}}}$ denote the gradient vectors of the individual loss terms, as defined in Section~\ref{sec:loss}, at training step $s$. The average normalized gradient is computed as

\begin{equation}
    \bar{g}^{(s)}
    =
    \frac{1}{\lvert \mathcal{J}_{\mathcal{L}} \rvert}
    \sum_{\ell \in \mathcal{J}_{\mathcal{L}}}
    \frac{g_\ell^{(s)}}{\left\lVert g_\ell^{(s)} \right\rVert},
    \label{eq:average_normalized_gradient}
\end{equation}

where $\lvert \mathcal{J}_{\mathcal{L}} \rvert$ is the number of loss terms. The gradient alignment score is then defined as

\begin{equation}
    \mathcal{A}^{(s)}
    =
    2\left\lVert \bar{g}^{(s)} \right\rVert^2 - 1.
    \label{eq:gradient_alignment_score}
\end{equation}

The score takes values in $[-1,1]$, with larger values indicating stronger directional agreement among the loss gradients. When $\lvert \mathcal{J}_{\mathcal{L}} \rvert = 2$, it reduces to the standard cosine similarity between the two normalized gradients.

\begin{figure}[h]
	\centering
	\includegraphics{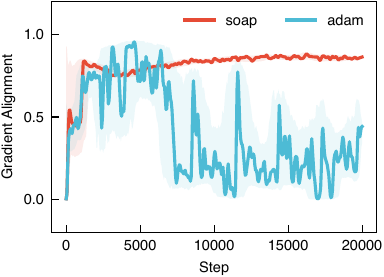}
	\caption{Gradient alignment score~\cite{wangGradientAlignmentPhysicsinformed2025a} between different loss components during training on the shallow water benchmark with PI-DeepONet. Adam's alignment collapses after approximately \num{5000} steps and remains near zero, indicating severe directional gradient conflicts on this multi-objective physics-informed loss. SOAP, by operating in a preconditioned eigenbasis, maintains a consistently high alignment score of approximately \num{0.90}, validating its suitability for this scenario.}
	\label{fig:gradient_alignment}
\end{figure}

Recent studies~\cite{wangGradientAlignmentPhysicsinformed2025a} suggest that quasi-second-order optimizers, by incorporating preconditioning and curvature information, are more effective at navigating such conflicting gradient landscapes. We therefore adopt SOAP~\cite{vyas2024soap} as our default optimizer. This behavior is illustrated in Figure~\ref{fig:gradient_alignment} using PI-DeepONet on the shallow water benchmark, where multiple coupled objectives induce pronounced directional conflicts; SOAP achieves substantially higher gradient alignment during training compared to Adam.

SOAP combines Shampoo-style preconditioning with Adam-style adaptive updates~\cite{vyas2024soap}. The preconditioner is constructed from the eigenvectors of the running gradient second-moment matrices:
\begin{equation}
	Q_L = \mathrm{eig}\!\left(\mathbb{E}[G^{(s)} G^{(s)\top}]\right), \qquad Q_R = \mathrm{eig}\!\left(\mathbb{E}[G^{(s)\top} G^{(s)}]\right),
\end{equation}
where $G^{(s)}$ denotes the raw gradient matrix at training step $s$. The gradient is then projected into the eigenbasis defined by $Q_L$ and $Q_R$, and the parameter update is Adam applied in the rotated space:
\begin{equation}
	\Delta\widetilde{W} = \eta\;\mathrm{Adam}\!\left(Q_L^\top G^{(s)}\, Q_R\right).
\end{equation}
By preconditioning the total gradient in an empirical curvature eigenbasis, SOAP can rescale poorly conditioned directions and reduce the adverse effect of conflicting component gradients on the final update. We report the performance of both Adam and SOAP in the ablation study in Section~\ref{sec:ablation_optimizer}, demonstrating that SOAP generally improves the predictive accuracy of PINOs for most of the considered problems.

\subsection{Training data and collocation sampling strategies}
\label{sec:sampling_strategies}

In physics-informed operator learning, sampling plays a role beyond constructing minibatches. Unlike supervised operator learning, where training signals are restricted to a finite set of labeled input--output trajectories, the physics-informed objective can be evaluated at arbitrary input functions and arbitrary points in the spatiotemporal domain. This provides an additional source of supervision: the model can be constrained by the governing equations not only at locations where reference solutions are available, but also throughout the broader parametric and physical domain. The choice of how input functions, collocation points, and labeled trajectories are sampled therefore directly affects the coverage of the training objective and the generalization behavior of the learned operator.

At each training step, we sample a batch of input functions $\{u_j\}_{j=1}^M$ from the parametric input space $\mathcal{U}$. For coordinate-based models, we then sample shared sets of query coordinates for the residual, initial-condition, and boundary-condition losses. Specifically, residual collocation points are sampled from the interior of the spatiotemporal domain, initial-condition points from the initial-time slice, and boundary-condition points from the spatial boundary:
\begin{align}
    \{u_j\}_{j=1}^{M}
        &\sim \mathcal{U}, \\
    \{(\mathbf{x}_i, t_i)\}_{i=1}^{N_r}
        &\overset{\text{LHS}}{\sim} \Omega \times (0, T], \\
    \{(\mathbf{x}_i, 0)\}_{i=1}^{N_{\text{ic}}}
        &\overset{\text{LHS}}{\sim} \Omega, \\
    \{(\mathbf{x}_i, t_i)\}_{i=1}^{N_{\text{bc}}}
        &\overset{\text{LHS}}{\sim} \partial\Omega \times [0, T].
\end{align}
Here Latin hypercube sampling is used to promote broad coverage of the sampling domain at each iteration. Resampling both the input functions and the collocation points during training allows the physics loss to probe a much larger portion of $\mathcal{U}\times\Omega\times[0,T]$ than would be possible with a fixed finite dataset.

\begin{figure}[h]
	\centering
	\includegraphics[width=\textwidth]{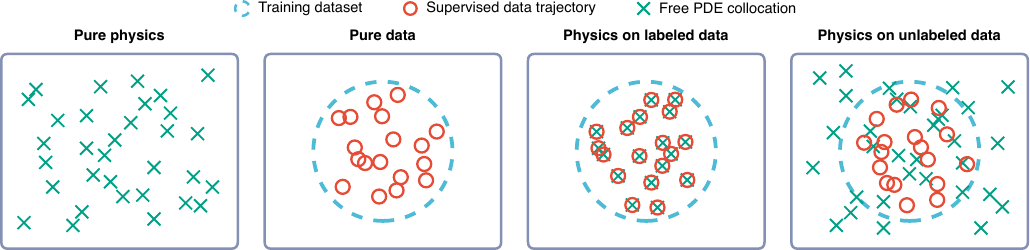}
	\caption{Illustration of the four training regimes defined by the use of supervised data and by where the physics residual is evaluated. Each square represents the full parametric and spatiotemporal domain $\mathcal{U}\times \Omega \times [0, T]$, while the dashed circles indicate the finite labeled dataset.}
	\label{fig:sampling_regimes}
\end{figure}

This distinction becomes important when comparing physics-informed and data-driven training. The presence of a PDE residual loss alone does not specify where the residual is evaluated. If the residual is evaluated only on the coordinates contained in labeled trajectories, then the physics loss inherits the same coverage limitations as the dataset. In contrast, if the residual is evaluated at freely sampled collocation points, the PDE provides supervision beyond the labeled data distribution. To separate these effects, we consider four training regimes, illustrated in Figure~\ref{fig:sampling_regimes}:
\begin{itemize}[leftmargin=*]
	\item \textbf{Pure physics}: the model is trained using only PDE residual, initial-condition, and boundary-condition losses, with both input functions and collocation points freely sampled during training;
	\item \textbf{Pure data}: the model is trained using the supervised data loss on labeled trajectories, with optional initial- and boundary-condition losses applied in the same way as in the physics-informed setting;
	\item \textbf{Physics on labeled data}: supervised and physics losses are both used, but the PDE residual is evaluated only at the spatiotemporal coordinates contained in the labeled trajectories;
	\item \textbf{Physics on unlabeled collocation points}: the supervised data loss is evaluated on labeled trajectories, while the physics losses are evaluated at freely sampled collocation points independent of the labeled dataset.
\end{itemize}

This categorization isolates the effect of free collocation sampling from the effect of adding labeled data. As shown in Section~\ref{sec:ablation_data}, this distinction is crucial: a well-resolved purely physics-informed training procedure with freely sampled collocation points can match or outperform data-driven training across several data regimes, whereas restricting the residual to labeled trajectory coordinates substantially weakens the benefit of the physics loss.

\subsection{Practical tips}
\label{sec:practical_tips}

Beyond the core training components discussed above, several implementation details also have a noticeable impact on stability and final accuracy in practice. In this subsection, we summarize a few empirically useful design choices that proved helpful for PINO training in our experiments.

\subsubsection{Enforcing periodic boundary conditions}

For PDEs with periodic spatial boundary conditions, soft enforcement via loss terms yields only approximate satisfaction and introduces additional hyperparameters. We instead hard-encode periodicity by replacing the raw spatial coordinates with a periodic embedding before they are passed to the decoder. For a 2D domain with periods $L_x$ and $L_y$, the embedding is defined as
\begin{equation}
	\Phi(\mathbf{x}) = \left[\sin\!\left(\frac{2\pi}{L_x}x_1\right),\; \cos\!\left(\frac{2\pi}{L_x}x_1\right),\; \sin\!\left(\frac{2\pi}{L_y}x_2\right),\; \cos\!\left(\frac{2\pi}{L_y}x_2\right)\right]. \label{eq:periodic_embedding}
\end{equation}
Since $\Phi(\mathbf{x})$ is periodic by construction, the decoded solution is guaranteed to satisfy the periodic boundary conditions exactly, without the need for additional loss terms and corresponding sampling.

\subsubsection{FiLM layer for time conditioning}

In the standard decoder, spatial and temporal Fourier embeddings are concatenated to form the initial query features, treating space and time symmetrically. This is suboptimal for PDEs in which the morphological evolution of the solution is predominantly driven by time, with the spatial pattern serving as a consistent template that is globally modulated across time steps---as in the Allen--Cahn-governed ice melting problem, where temporal dynamics primarily govern the displacement of the interface, or the wave equation, where waveforms propagate while their shape is dictated by the spatial structure.

For such problems, we adopt Feature-wise Linear Modulation (FiLM)~\cite{brockschmidt2020gnn} to decouple spatial and temporal processing in the decoder. The spatial Fourier embedding $\mathbf{q}_x$ alone serves as the initial query, such that
\begin{equation}
	\mathbf{q}_0 = \mathbf{q}_x(\mathbf{x}) \in \mathbb{R}^{N_q\times D_e},
\end{equation}
while the temporal embedding $\mathbf{q}_t$ is passed through a shared MLP to produce a layer-wise scale and shift:
\begin{equation}
	[\boldsymbol{\gamma}^{(l)},\, \boldsymbol{\beta}^{(l)}] = \mathrm{MLP}_{\mathrm{FiLM}}(\mathbf{q}_t) \in \mathbb{R}^{2D_d}, \quad l = 1, \ldots, L_d.
\end{equation}
At each decoder layer $l$, the normalized query features are modulated before cross-attention:
\begin{equation}
	\tilde{\mathbf{q}}^{(l)} = \mathrm{LN}(\mathbf{q}^{(l)}) \odot \bigl(1 + \boldsymbol{\gamma}^{(l)}\bigr) + \boldsymbol{\beta}^{(l)}.
\end{equation}
To ensure training stability, the last linear layer of $\mathrm{MLP}_{\mathrm{FiLM}}$ is initialized to zero, so the modulation starts as the identity at the beginning of training. We find FiLM beneficial for problems with the above structure (ice melting and wave equation) but not for cases with more intricate spatial-temporal coupling such as Burgers' equation or the shallow water equations. See Section~\ref{sec:ablation_film} for an ablation study.

\subsubsection{Warm-up phase of training}

A key advantage of physics-informed neural operators is their ability to draw a fresh batch of collocation points at every training step, effectively enabling infinite sampling of the function space and the spatiotemporal domain. In practice, we resample both the input functions $\{u_j\}$ and the query coordinates $\{(\mathbf{x}_i, t_i)\}$ at every gradient step to fully exploit this property. However, per-step resampling introduces high variance in the gradient across consecutive iterations, which prevents the initial condition loss from converging: the optimizer cannot accumulate a consistent signal to anchor the solution at $t = 0$, weakening the causal structure of the training dynamics, and the network is prone to collapsing to the trivial zero solution~\cite{krishnapriyan2021characterizing}.

To mitigate this, we employ a warm-up training of \num{50}--\num{2000} steps in which we slow down the resampling frequency to every \num{50}--\num{100} steps. During this phase, the model is exposed to a more consistent set of input functions and collocation points, allowing the network to first acquire a rough approximation of the solution structure before being exposed to the full stochasticity of per-step resampling. We also up-weight the initial condition losses by an additional factor of \num{2}--\num{10} on top of the GradNorm weights, so that the optimizer is strongly guided toward satisfying the prescribed data at $t = 0$ before the PDE residual loss dominates. After the warm-up phase, the resampling frequency and loss weights are restored to their standard values.

\begin{table}[t]
	\centering
	\caption{Relative $L^2$ error (in \%, mean $\pm$ std) of physics-informed neural operators on five benchmark PDEs, evaluated over the full test set.}
	\label{tab:main_results}
	\begin{tblr}{
	width = \textwidth,
	colspec = {Q[c,1]Q[c,1]Q[c,1]Q[c,1,font=\bfseries]},
	row{1} = {font=\bfseries},
	}
	\hline
	Benchmark PDE & PI-DeepONet & PI-FNO & PI-CViT \\
	\hline
	Burgers & \qty{34.5 +- 2.54}{\percent} & \qty{9.67 +- 1.84}{\percent} & \qty{0.78 +- 0.11}{\percent} \\
	Wave & \qty{11.9 +- 1.82}{\percent} & \qty{19.1 +- 2.29}{\percent} & \qty{2.46 +- 0.45}{\percent} \\
	Shallow Water & \qty{29.5 +- 3.77}{\percent} & \qty{9.02 +- 0.97}{\percent} & \qty{4.29 +- 0.61}{\percent} \\
	Ice melting & \qty{37.1 +- 9.35}{\percent} & \qty{52.9 +- 1.00}{\percent} & \qty{1.87 +- 0.68}{\percent} \\
	Lid-driven Cavity & \qty{7.60 +- 1.75}{\percent} & \qty{103.5 +- 2.67}{\percent} & \qty{6.46 +- 0.57}{\percent} \\
	\hline
	\end{tblr}
\end{table}

\begin{figure}[t]
	\centering
	\includegraphics[width=\textwidth]{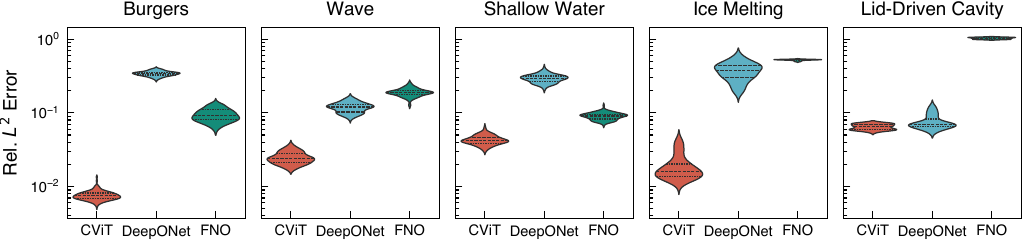}
	\caption{Relative $L^2$ error distributions across all benchmark PDEs. Each violin shows the full distribution over 100 test samples. PI-CViT consistently achieves lower errors and tighter distributions compared to PI-DeepONet and PI-FNO.}
	\label{fig:error_distribution}
\end{figure}

\section{Results}
\label{sec:results}
We now evaluate whether the training principles developed above lead to accurate and robust physics-informed operator learning in practice. The goal is not only to compare architectures, but also to test whether a fully physics-informed pipeline can learn solution operators across diverse PDE families without relying on supervised solution data.

We compare PI-CViT against its physics-informed counterparts built on DeepONet and FNO (referred to as PI-DeepONet and PI-FNO, respectively) across these five benchmarks. All three models are trained with a pure physics loss without any supervised data. Unless otherwise stated in the description of each benchmark, we follow the standard training pipeline described in Section~\ref{sec:method}, including GradNorm weighting, causal weighting, and the SOAP optimizer. PI-CViT and PI-DeepONet evaluate PDE residuals at randomly sampled input functions and spatiotemporal collocation points, with derivatives computed via automatic differentiation; PI-FNO predicts the full solution trajectory on a fixed spatiotemporal grid in a single forward pass and computes residuals via spectral/finite-difference differentiation combined with a fourth-order Runge--Kutta (RK4) time integrator. For each benchmark, the trained models of the three architectures are evaluated on a held-out test set of 100 trajectories with unseen input functions. Further details on hyperparameter settings, architectural configurations, and test set generation are provided in Appendices~\ref{app:numerical_solvers} and~\ref{app:hyperparameters}.

Table~\ref{tab:main_results} reports mean and standard deviation of the relative $L^2$ error (see Appendix~\ref{app:metric} for the definition) over the full test set for each benchmark, and Figure~\ref{fig:error_distribution} shows the full error distributions. PI-CViT achieves the lowest mean error and the tightest distribution across all benchmarks. The gain is most pronounced on Burgers' equation, where PI-CViT outperforms PI-FNO by over an order of magnitude.
On the lid-driven cavity benchmark, PI-DeepONet remains competitive since this problem is parameterized solely by a scalar Reynolds number, leaving less room for the structured input encoding of PI-CViT to demonstrate its advantages. Nevertheless, the distribution plots reveal that PI-DeepONet's errors are significantly more dispersed, with a heavier tail and a larger maximum error; these failure cases are concentrated at high Reynolds numbers, where the flow develops stronger nonlinearities. PI-FNO performs poorly on this benchmark, as its spectral convolution assumes periodic boundary conditions that are incompatible with the no-slip walls and corner singularities of the lid-driven cavity geometry. 

The subsequent sections present the problem setup and detailed solution comparisons for each benchmark PDE. A comprehensive ablation study on the training pipeline components is provided in Section~\ref{sec:ablation}, and additional results and visualizations are included in Appendix~\ref{app:additional_visualizations}.


\subsection{Burgers' equation} 
\label{sec:burgers}
We begin with the 2D vector Burgers' equation, a classical benchmark for nonlinear advection--diffusion dynamics, governing the evolution of a velocity field $\mathbf{v}(\mathbf{x}, t) = (v_1, v_2)$ under the combined effects of nonlinear advection and viscous diffusion:
\begin{equation}
	\frac{\partial \mathbf{v}}{\partial t} + (\mathbf{v} \cdot \nabla) \mathbf{v} = \nu \nabla^2 \mathbf{v},
\end{equation}
where $\nu = 0.01$ is the kinematic viscosity. Periodic boundary conditions are imposed in both spatial dimensions. The parametric family consists of random initial velocity fields, drawn from a periodic Gaussian random field (GRF) with a fixed Gaussian power spectrum (length scale $\ell$ and amplitude $A$; see Appendix~\ref{app:grf}), with each velocity component sampled independently. The neural operator learns to map the initial velocity field $\mathbf{v}(\mathbf{x}, 0)$ to the velocity solution $\mathbf{v}(\mathbf{x}, t)$ for $t \in [0, 1]$.

\begin{figure}[t]
	\centering
	\includegraphics[width=0.95\textwidth]{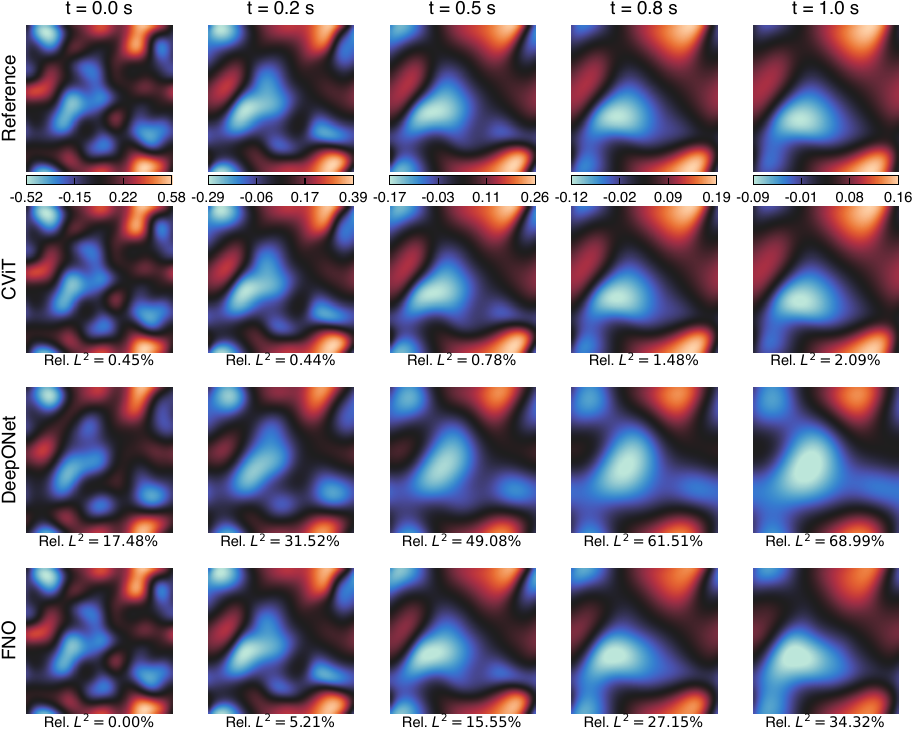}
	\caption{Burgers' equation. The first velocity component $v_1$ predicted by different physics-informed neural operators compared with the reference solution. The second velocity component exhibits qualitatively similar behavior and is omitted for brevity.}
	\label{fig:burgers_solution_all_models}
\end{figure}

Figure~\ref{fig:burgers_solution_all_models} shows the first velocity component $v_1$ at five time snapshots for a representative test sample. Since the two velocity components exhibit qualitatively similar behavior, we visualize only $v_1$ here. PI-CViT closely tracks the reference solution throughout the entire time window, with the maximum error of \qty{2.09}{\percent} at the final snapshot $t=1.0$. PI-DeepONet exhibits rapid error accumulation: while it partially captures the large-scale flow structure at early times, the prediction progressively deteriorates and by $t=1.0$ the relative $L^2$ error reaches nearly \qty{70}{\percent}, indicating that the model struggles to sustain accurate temporal dynamics under the nonlinear advection. PI-FNO takes the $t=0$ snapshot directly as input and predicts the full spatiotemporal solution trajectory in a single forward pass, so its error at $t=0$ is zero by construction; nevertheless, it accumulates approximately \qty{34}{\percent} error by the end of the time horizon, retaining the dominant spatial structures but losing fine-scale detail at later times.

\subsection{Wave equation}
\label{sec:wave}

We consider the 2D scalar wave equation with a spatially varying wave speed $c(\mathbf{x})$, which serves as a challenging benchmark for heterogeneous wave propagation:
\begin{align}
	\frac{\partial^2 v}{\partial t^2} &= c(\mathbf{x})^2 \nabla^2 v, \quad \mathbf{x} \in [0,1)^2, \quad t \in [0,1], \\
	c(\mathbf{x}) &= 1 + \tfrac{1}{2}\sin(2\pi x_1)\sin(2\pi x_2).
\end{align}
The initial displacement is drawn from a periodic GRF with a Gaussian power spectrum in Eq.~\eqref{eq:grf}, and the initial velocity is set to zero:
\begin{align}
	v(\mathbf{x}, 0) &= \xi(\mathbf{x}), \\
	\frac{\partial v}{\partial t}(\mathbf{x}, 0) &= 0.
\end{align}
Periodic boundary conditions are imposed in both spatial dimensions. The solution operator takes the initial displacement field $v(\mathbf{x}, 0)$ as input and outputs the wave solution $v(\mathbf{x}, t)$ for $t \in [0, 1]$.

\begin{figure}[t]
	\centering
	\includegraphics[width=0.95\textwidth]{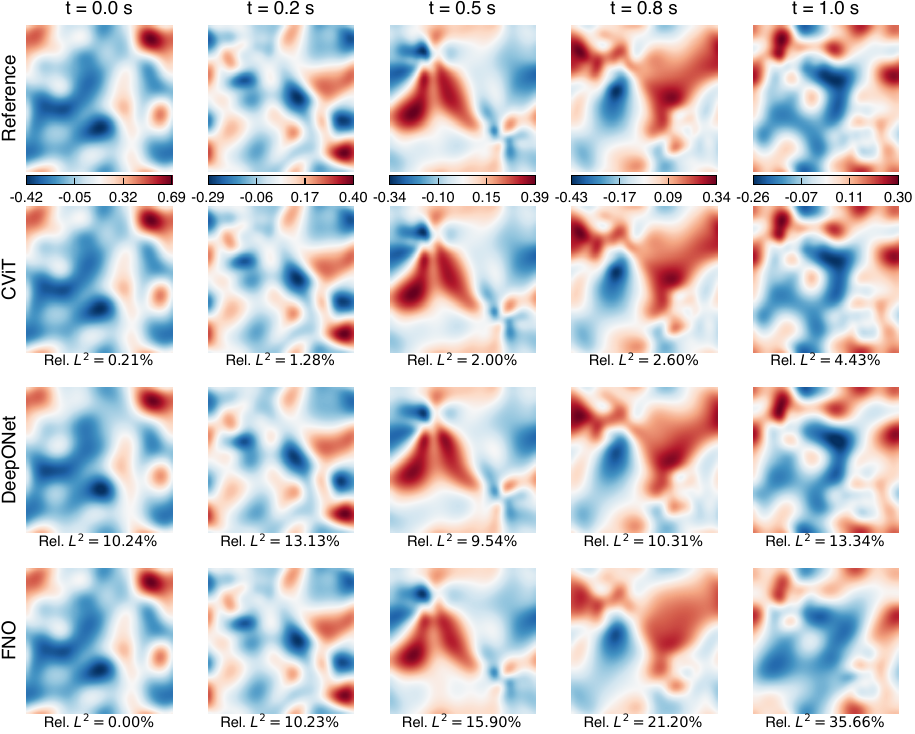}
	\caption{Wave equation. Scalar solution field $v$ predicted by different physics-informed neural operators compared with the reference solution.}
	\label{fig:wave_solution_all_models}
\end{figure}

Figure~\ref{fig:wave_solution_all_models} shows the predicted displacement field at five time snapshots. PI-CViT accurately tracks the wave field throughout, with errors growing gradually from \qty{0.21}{\percent} at $t=0$ to \qty{4.43}{\percent} at $t=1.0$, outperforming the other two models by a wide margin. PI-DeepONet exhibits uniformly high errors of around \qty{10}{\percent} across all time snapshots.
PI-FNO starts at zero error by construction and accumulates monotonically to \qty{35.66}{\percent} by $t=1.0$, with the predicted field losing spatial coherence at later times.

\begin{figure}[t]
	\centering
	\includegraphics[width=0.95\textwidth]{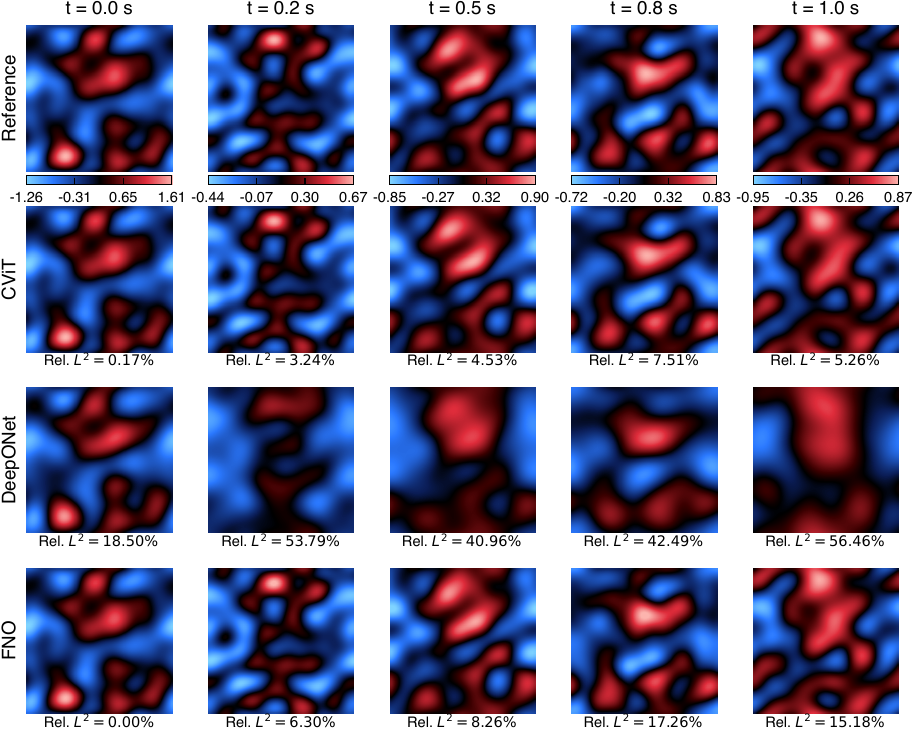}
	\caption{Shallow water equations. Free-surface height $h$ predicted by different physics-informed neural operators compared with the reference solution.}
	\label{fig:swe_solution_all_models}
\end{figure}

\subsection{Shallow water equations}
\label{sec:shallow_water}

The linearized shallow water equations govern the evolution of a fluid layer with free-surface height $h$ and depth-averaged horizontal velocity $\mathbf{v} = (v_1, v_2)$ under gravity and the Coriolis effect:
\begin{align}
	\frac{\partial h}{\partial t} + H \nabla \cdot \mathbf{v} &= 0, \\
	\frac{\partial \mathbf{v}}{\partial t} + g \nabla h + f \mathbf{v}^\perp &= 0,
\end{align}
where $H$ is the mean fluid depth, $g$ is the gravitational acceleration, $f$ is the Coriolis parameter, and $\mathbf{v}^\perp = (-v_2, v_1)$ denotes the $90^\circ$ rotation of $\mathbf{v}$. We set $g = H = 1$ and use a large Coriolis parameter $f = 10$ to drive strong inertia-gravity wave activity, with a characteristic oscillation period $2\pi/f \approx 0.63$ comparable to the time horizon $T = 1$. The fluid height is initialized with a perturbation drawn from a periodic GRF (see Appendix~\ref{app:grf}), while the velocity field is initialized to zero:
\begin{align}
	h(\mathbf{x}, 0) &= \xi(\mathbf{x}), \\
	\mathbf{v}(\mathbf{x}, 0) &= \mathbf{0}.
\end{align}
Periodic boundary conditions are imposed in both spatial dimensions. The solution operator maps the initial height perturbation $h(\mathbf{x}, 0)$ to the full coupled state $(h, \mathbf{v})(\mathbf{x}, t)$ for $t \in [0, 1]$. The primary challenge of this benchmark lies in the tight coupling among the three output fields $(h, v_1, v_2)$ and the dispersive inertia-gravity wave dynamics induced by the large Coriolis parameter, which drive rapid geostrophic adjustment and fast multi-scale oscillations throughout the time window.

Figure~\ref{fig:swe_solution_all_models} shows the predicted height field $h$ at five time snapshots. The reference solution undergoes rapid geostrophic adjustment immediately after $t=0$, with the field amplitude dropping sharply before inertia-gravity waves re-emerge and the solution oscillates at later times. PI-CViT follows this evolution closely, with the final error remaining below \qty{6}{\percent}. PI-DeepONet struggles at early times, reaching over \qty{50}{\percent} error at $t=0.2$ and remaining above \qty{40}{\percent} thereafter, and the predicted field is visibly over-smoothed. PI-FNO performs considerably better than PI-DeepONet, with errors growing from zero to approximately \qty{10}{\percent} by $t=1.0$. The strong multi-field coupling and fast oscillations make PI-CViT's accuracy on this benchmark particularly notable.

\begin{figure}[t]
	\centering
	\includegraphics[width=0.95\textwidth]{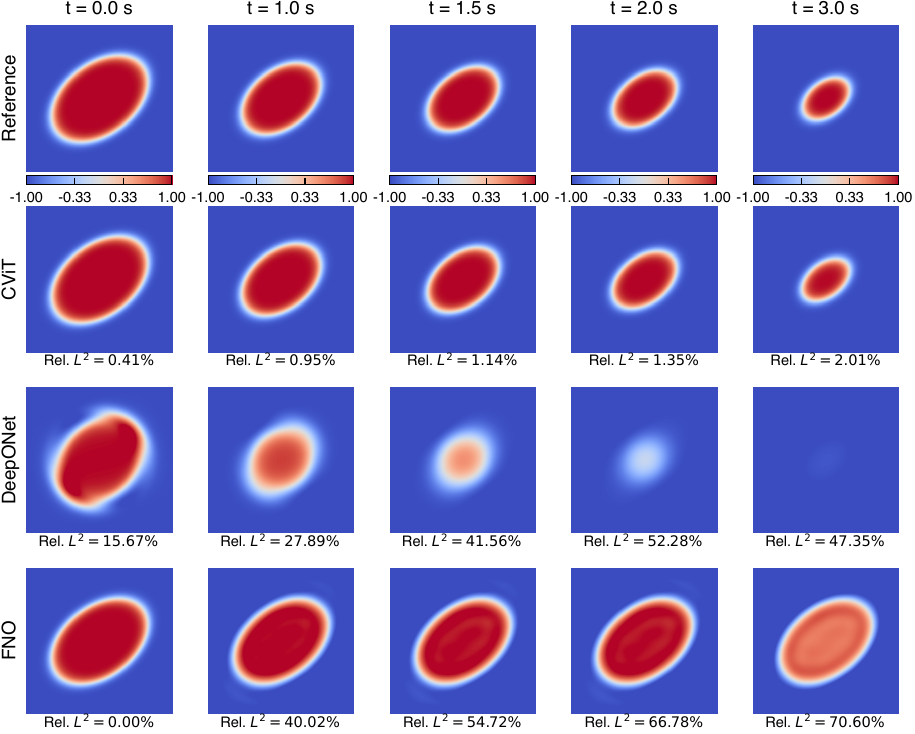}
	\caption{Ice melting problem. Phase-field variable $\phi$ predicted by different physics-informed neural operators compared with the reference solution.}
	\label{fig:ice_melting_solution_all_models}
\end{figure}

\subsection{Ice melting}
\label{sec:ice_melting}

Unlike the previous benchmarks whose initial conditions are smooth random fields, the ice melting problem~\cite{jianwangPhaseFieldModelingNumerical2021} is defined by a structured parametric family of elliptical initial inclusions, each with a well-defined thin interface separating the two phases and varying in shape and orientation. The resulting operator-learning task must therefore resolve a thin moving interface governed by stiff phase-field dynamics while remaining robust to variations in the embedded elliptical inclusion. A 2D Allen--Cahn phase-field model governs the ice--water phase transition, where the order parameter $\phi(\mathbf{x}, t) \in [-1, 1]$ transitions smoothly between the ice phase ($\phi = +1$) and the water phase ($\phi = -1$) through a diffuse interface of width $O(\epsilon)$:
\begin{equation}
	\frac{\partial \phi}{\partial t} = M \left( \nabla^2 \phi - \frac{F'(\phi)}{\epsilon^2} \right) - \lambda \frac{\sqrt{2F(\phi)}}{\epsilon},
\end{equation}
where $M=0.1$ is the interface mobility, $\lambda=5$ is a latent-heat coupling coefficient, and $F(\phi) = \frac{1}{4}(\phi^2 - 1)^2$ is the double-well potential. The interface thickness is determined by $\epsilon={3h}/(\sqrt{2}\mathrm{arctanh}(0.9))$, where $h=100/63$ is the grid spacing of the reference solution (see Appendix~\ref{app:numerical_solvers}).
The initial condition is prescribed as a diffuse-interface profile of a randomly sampled rotated elliptical inclusion:
\begin{equation}
	\phi(\mathbf{x}, 0) = \tanh\!\left(\frac{d(\mathbf{x})}{\sqrt{2}\,\epsilon}\right),
\end{equation}
where $d(\mathbf{x})$ is an approximate signed distance to the ellipse boundary; see Appendix~\ref{app:ellipse} for details. The problem is posed on a domain of $\Omega=[-50, 50]^2$ with homogeneous Neumann boundary conditions. Given the initial phase-field profile $\phi(\mathbf{x}, 0)$ parameterized by the ellipse geometry $(a, b, \theta)$, the neural operator predicts the phase-field evolution $\phi(\mathbf{x}, t)$ for $t \in [0, 3]$. For this benchmark, we use Adam optimizer instead of SOAP, as it yields better performance on this phase-field problem; see Appendix~\ref{app:negative_results} for a detailed analysis.

Figure~\ref{fig:ice_melting_solution_all_models} shows the phase-field variable $\phi$ for a representative elliptical inclusion. The reference solution exhibits a steadily shrinking ellipse with a thin, well-defined interface throughout. PI-CViT accurately captures the interface position and shape at all time steps, with errors growing only from \qty{0.41}{\percent} at $t=0$ to \qty{2.01}{\percent} at $t=3.0$. PI-DeepONet and PI-FNO both perform noticeably worse on this benchmark. PI-DeepONet already incurs \qty{15.67}{\percent} error at $t=0$, suggesting that its fully-connected branch network has difficulty encoding the thin diffuse interface from the flattened input field; the predicted interface progressively loses its sharp profile, with the elliptical shape becoming indistinct by $t=2.0$. PI-FNO, despite taking the exact initial condition as input, degrades substantially over time and reaches \qty{70.60}{\percent} error by $t=3.0$, with unphysical artifacts appearing in the interior at later times. This indicates that PI-FNO does not capture the physical dynamics of the phase-field evolution. This behavior is likely due to the limited ability of truncated spectral convolutions combined with finite-difference residuals to resolve the thin interface and the stiff nonlinearity of the double-well potential term.

\begin{figure}[t]
	\centering
	\includegraphics[width=0.95\textwidth]{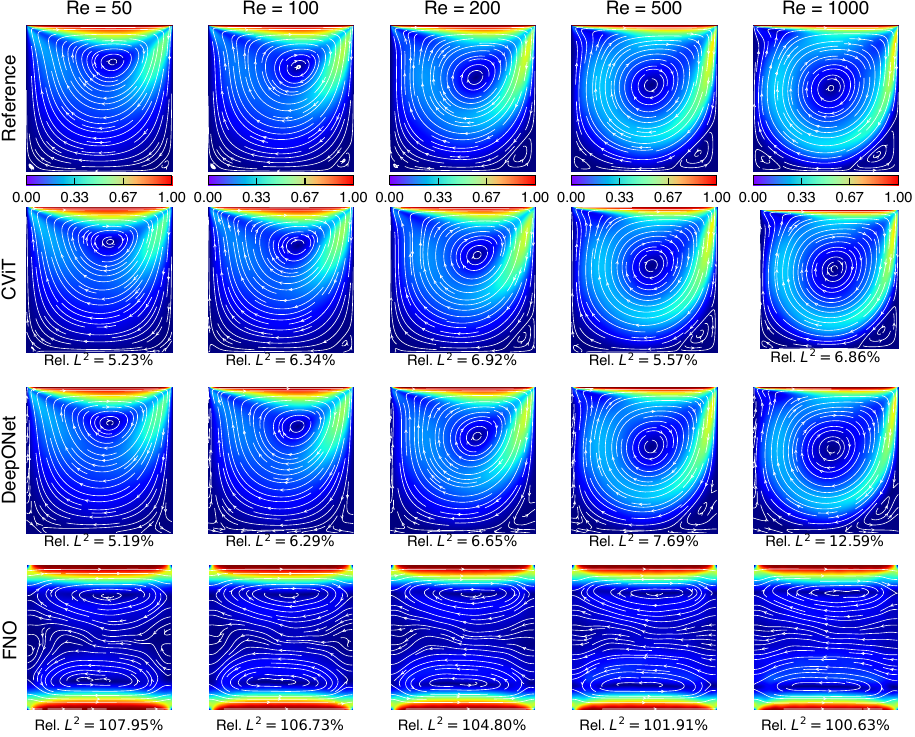}
	\caption{Lid-driven cavity flow. Streamlines and velocity magnitude predicted by different physics-informed neural operators compared with the reference solution.}
	\label{fig:ldc_solution_all_models}
\end{figure}

\subsection{Lid-driven Cavity flow}
\label{sec:lid_driven_cavity}

The lid-driven cavity flow is a classical steady-state benchmark for the incompressible Navier--Stokes equations with complex boundary conditions and corner singularities. The governing equations are given by
\begin{align}
	(\mathbf{v} \cdot \nabla) \mathbf{v} &= -\nabla p + \nu \nabla^2 \mathbf{v}, \\
	\nabla \cdot \mathbf{v} &= 0,
\end{align}
where $\mathbf{v} = (v_1, v_2)$ is the velocity field, $p$ is the pressure, and $\nu = 1/\mathrm{Re}$ is the kinematic viscosity. No-slip conditions are imposed on the bottom, left, and right walls, while the top boundary moves with unit tangential velocity:
\begin{equation}
	\mathbf{v}(\mathbf{x}) = 
	\begin{cases}
		\mathbf{0}, & \text{for stationary walls } (x_2 = 0,\; x_1 = 0,\; x_1 = 1), \\[4pt]
		(1, 0), & \text{for moving lid } (x_2 = 1).
	\end{cases}
\end{equation}
To avoid the velocity discontinuity at the top corners, the lid velocity is replaced by a smooth approximation:
\begin{equation}
	v_1(x_1,1)=1-\frac{\cosh(50(x_1-0.5))}{\cosh(25)},
\qquad
v_2(x_1,1)=0.
\end{equation}
The Reynolds number is the sole problem parameter, sampled log-uniformly over $\mathrm{Re} \in [50, 1000]$; see Appendix~\ref{app:re} for details. Unlike the previous benchmarks, the solution operator here maps a scalar $\mathrm{Re}$ directly to the steady-state velocity and pressure fields $(\mathbf{v}, p)(\mathbf{x})$. As $\mathrm{Re}$ increases, the primary vortex center shifts and secondary corner vortices develop, producing qualitatively distinct flow patterns across the parameter range. Since this benchmark is steady-state and has no temporal dimension, causal weighting is not applied, whereas GradNorm weighting is still used to balance the different physics loss terms. For PI-FNO, the spatial domain is padded by 32 grid points on each side to handle the non-periodic walls (see Appendix~\ref{app:hyperparameters}).

Figure~\ref{fig:ldc_solution_all_models} shows the predicted streamlines and velocity magnitude at five representative Reynolds numbers. PI-CViT maintains consistent accuracy across the entire parameter range, with errors between \qty{5.23}{\percent} and \qty{6.86}{\percent}. PI-DeepONet achieves comparable performance at low Reynolds numbers (e.g., \qty{5.19}{\percent} at $\mathrm{Re}=50$), but its accuracy degrades steadily as $\mathrm{Re}$ increases, reaching \qty{12.59}{\percent} at $\mathrm{Re}=1000$ where stronger nonlinearities and finer secondary vortex structures develop. Although we pad the spatial domain to mitigate the non-periodicity and corner singularities, PI-FNO struggles across all Reynolds numbers, with errors exceeding \qty{100}{\percent} throughout; the predicted velocity field near the bottom boundary departs sharply from the reference, reflecting the difficulty of imposing non-periodic no-slip and lid boundary conditions within a spectral convolution architecture.

\subsection{Ablation studies}
\label{sec:ablation}
In this section, we conduct  ablation studies of the main components of the proposed physics-informed training pipeline. For each study, only the target component is varied, while all other hyperparameters, architectural settings, sampling strategies, and training procedures are kept identical to the default pipeline. This allows us to isolate the effects of optimizer selection, labeled data usage, weighting schemes, derivative computation, and FiLM conditioning on accuracy and training behavior across representative benchmarks.

\begin{figure}[t]
	\centering
	\includegraphics[width=0.85\textwidth]{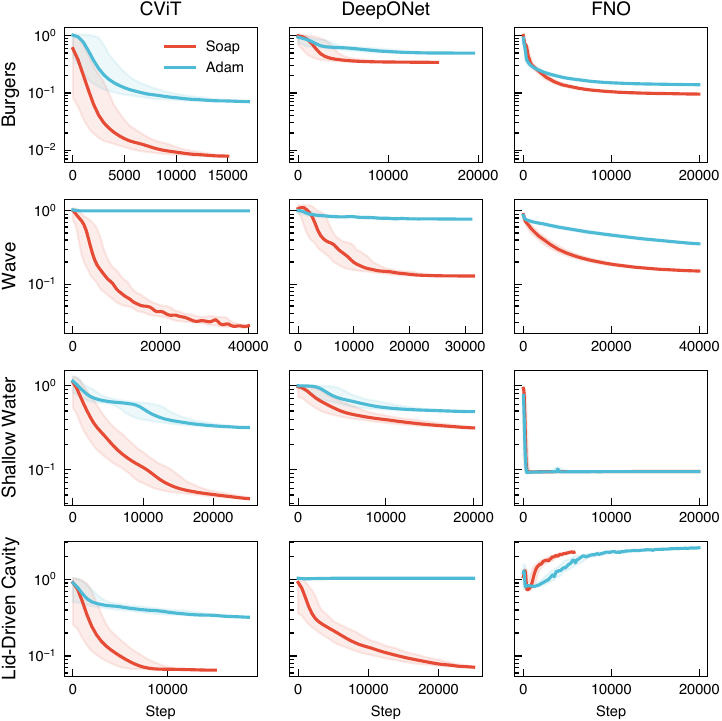}
	\caption{Ablation study on optimizer selection. Test error convergence curves during training for SOAP and Adam across three architectures and four benchmark PDEs.}
	\label{fig:ablation_optimizer}
\end{figure}

\paragraph{Optimizer selection.}
\label{sec:ablation_optimizer}
We compare SOAP against Adam across all three architectures and four benchmark PDEs; test error convergence curves during training are shown in Figure~\ref{fig:ablation_optimizer}.
PI-CViT achieves the lowest errors across all benchmarks with the SOAP optimizer, further corroborating Table~\ref{tab:main_results}.
More importantly, SOAP consistently outperforms Adam for both PI-CViT and PI-DeepONet, with the most pronounced gains on the Burgers and wave benchmarks, where the final error for PI-CViT is reduced by over an order of magnitude.
The ice melting benchmark is an exception where SOAP does not yield a clear improvement; we analyze this anomaly in Appendix~\ref{app:negative_results}.
For PI-FNO, SOAP outperforms Adam on the Burgers and wave benchmarks; however, on the shallow water benchmark, both optimizers plateau near \qty{10}{\percent} error throughout training, suggesting that the bottleneck lies in the architecture's limited spectral resolution and its finite-difference-based loss formulation for the multiscale oscillatory dynamics of this problem.
Finally, PI-FNO performs poorly on the lid-driven cavity benchmark under both optimizers, consistent with the difficulty of imposing non-periodic no-slip boundary conditions discussed in Section~\ref{sec:lid_driven_cavity}.

\paragraph{Labeled training data.}
\label{sec:ablation_data}

We investigate the role of labeled data by comparing four training regimes (as defined in Section~\ref{sec:sampling_strategies}) with varying numbers of labeled solution trajectories, using PI-CViT on the Burgers, shallow water, and lid-driven cavity benchmarks.
The labeled dataset is disjoint from the test set. Each trajectory is a full numerical solution on a \numproduct{64x64} spatial grid over \num{100} time snapshots for time-dependent problems. 
All data-driven regimes share the same batch size of input functions, total training steps, and full training pipeline (GradNorm weighting, causal weighting, and the SOAP optimizer) as our physics-informed training. The data loss is evaluated with batch size $M$, matching the number of input functions sampled per step in the physics-informed training. At each step, spatiotemporal coordinates are randomly subsampled from the labeled trajectories in the same quantity as $N_r$, and initial and boundary condition losses are applied identically.

\begin{figure}[h]
	\centering
	\includegraphics[width=\textwidth]{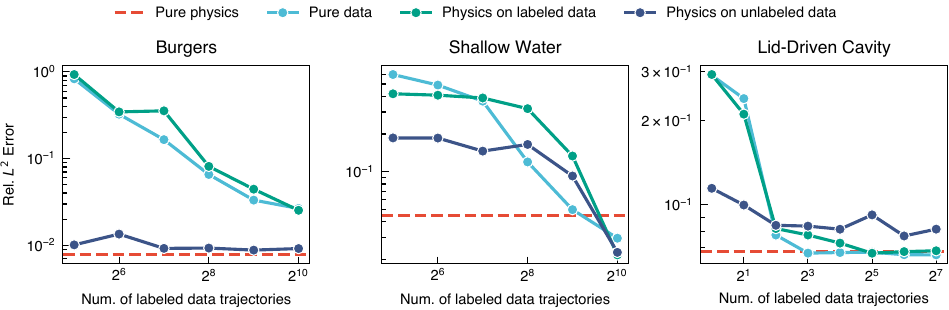}
	\caption{Ablation study on labeled training data. Relative $L^2$ error of four training regimes with varying numbers of labeled trajectories on Burgers, shallow water, and lid-driven cavity benchmarks. \textit{Pure physics} uses physics losses with freely sampled collocation and no labeled data. \textit{Pure data} uses supervised training only. \textit{Physics on labeled data} evaluates physics losses at the labeled trajectory coordinates. \textit{Physics on unlabeled data} combines freely sampled physics collocation with a supervised data loss. Pure physics (dashed red) matches or outperforms data-driven regimes across most of the data range, and adding labeled data does not significantly improve over pure physics, indicating that free collocation sampling is the critical ingredient and that gradient conflicts arise when physics and data losses are combined.}
	\label{fig:data_vs_physics}
\end{figure}

Several consistent patterns emerge from Figure~\ref{fig:data_vs_physics}.
Pure physics (dashed red) matches or outperforms data-driven regimes across most of the data range, demonstrating that a well-formulated physics loss with free collocation sampling can rival the accuracy of supervised training even at large labeled-data scales. 
We attribute this largely to free collocation sampling, as probing the full function space and spatiotemporal domain at every training step allows the physics loss to regularize the model, promoting generalization to unseen input functions and test coordinates while reducing the reliance on labeled data.
Pure data becomes competitive only with a large number of trajectories. For Burgers, a noticeable gap persists even at $2^{10}$ trajectories, and for the shallow water equations, approximate parity with pure physics requires around $2^{9}$ trajectories. This reflects the difficulty of learning a complex function-to-function mapping from a limited number of labeled solutions.
The lid-driven cavity is an exception where pure data converges quickly, requiring as few as $2^3$ trajectories. Unlike the other two benchmarks, which involve learning a function-to-function mapping from random initial fields, the lid-driven cavity operator maps a single scalar Reynolds number to the steady-state flow field. In this low-dimensional parameter space, the model need only interpolate among a small number of flow patterns, making pure data highly competitive with few labeled examples.

Restricting the physics loss to the labeled trajectory coordinates (physics on labeled data) underperforms both pure physics and pure data at every data scale on Burgers and shallow water. This confirms that free collocation sampling appears to be a critical factor in these experiments, since without coverage of the full spatiotemporal domain the physics loss cannot prevent the model from overfitting to the limited labeled trajectories.
Combining physics with a supervised data loss does not improve over pure physics and can occasionally degrade performance, as seen in the error bumps at $2^{7}$ trajectories on Burgers and $2^{8}$ trajectories on the shallow water equations. These degradations suggest that gradient conflicts between physics and data objectives may partially counteract their individual benefits, and developing training strategies that mitigate such conflicts warrants further investigation to unlock the potential of combining physics and data when the physics loss cannot be fully resolved.

\paragraph{Weighting schemes.}
\label{sec:ablation_weighting}

\begin{figure}[h]
	\centering
	\includegraphics[width=\textwidth]{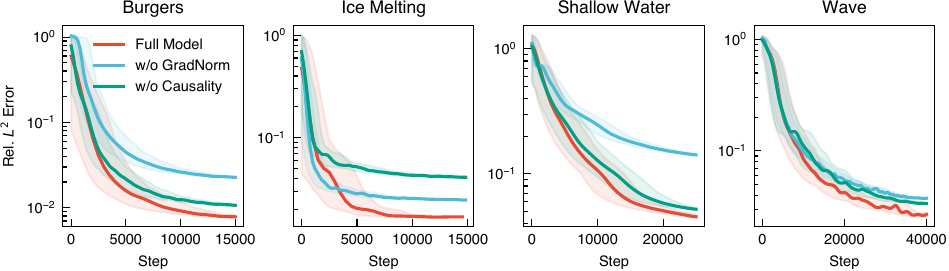}
	\caption{Ablation study on weighting schemes. Test error convergence curves during training for PI-CViT with and without GradNorm weighting and causal weighting on four benchmark PDEs. Full model with both weighting schemes consistently achieves the lowest final error.}
	\label{fig:ablation_gradnorm_causality}
\end{figure}

We ablate the two weighting schemes by removing each individually, using PI-CViT on four benchmarks; results are shown in Figure~\ref{fig:ablation_gradnorm_causality}.
The full model with both GradNorm and causal weighting consistently achieves the lowest final error, confirming the complementary benefits of these two schemes. 
Removing GradNorm is most harmful on the shallow water benchmark, where the gradient norm imbalance among the coupled loss terms is most severe (Figure~\ref{fig:gradnorms}).
Removing causal weighting most significantly degrades performance on the ice melting benchmark, where strict temporal ordering is critical for the optimizer to resolve the phase transition correctly.

\paragraph{Derivative and residual-evaluation strategy.}
\label{sec:ablation_derivative}

\begin{figure}[h]
	\centering
	\includegraphics{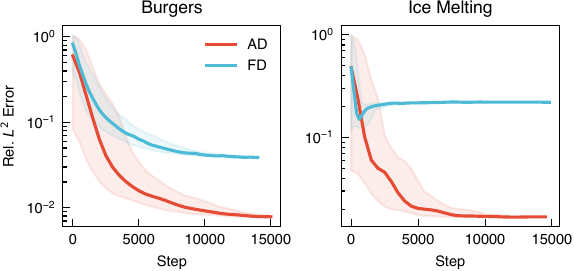}
	\caption{Ablation study on derivative computation method. Test error convergence curves during training for PI-CViT using automatic differentiation and finite-difference/spectral methods to compute PDE residuals on Burgers and ice melting benchmarks.}
	\label{fig:ablation_loss_fd_vs_ad}
\end{figure}

We compare automatic differentiation (AD) against finite-difference/spectral approximations (FD) for evaluating PDE residuals, using PI-CViT on Burgers and ice melting. Results are shown in Figure~\ref{fig:ablation_loss_fd_vs_ad}.
For FD, derivatives are computed on a fixed \numproduct{64x64} spatial grid with \num{100} uniform time steps using spectral differentiation (Burgers) or finite differences (ice melting) combined with an RK4 time integrator; AD evaluates residuals at freely sampled sparse collocation points via exact automatic differentiation.

\begin{figure}[h]
	\centering
	\subfigure{
		\includegraphics[width=0.54\textwidth]{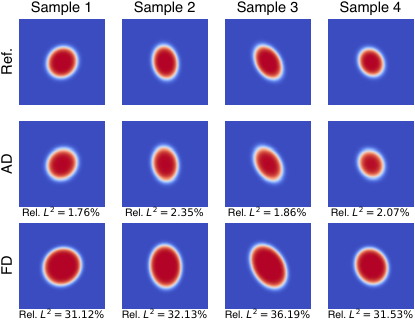}
	}
	\hfill
	\subfigure{
		\includegraphics[width=0.4\textwidth]{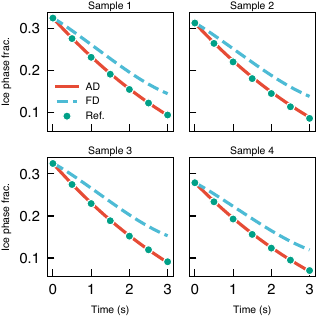}
	}
	\caption{FD versus AD on the ice melting benchmark. \textbf{Left}: predicted phase-field variable $\phi$ at $t = \qty{3}{\second}$ for four representative samples. FD produces spatially coherent solutions with clean interfaces, but the ellipses are systematically larger than the reference, indicating slower interface evolution. \textbf{Right}: ice phase fraction over time for the same samples. AD closely tracks the reference melting rate, while FD exhibits a consistent lag attributable to insufficient temporal resolution of the fixed discrete grid.}
	\label{fig:ablation_fd_vs_ad_icemelting}
\end{figure}

The AD-based free-collocation residual outperforms the fixed-grid FD/spectral residual on both benchmarks. Although FD avoids backpropagating through the network, it must evaluate the model at all \numproduct{64x64x100} grid points per step, incurring a wall-clock time of \qty{3.85}{h} per \num{10000} steps on ice melting and \qty{6.53}{h} on Burgers, compared to \qty{0.65}{h} and \qty{1.00}{h} for AD. Crucially, even this already costly dense discretization proves insufficient in accuracy. Figure~\ref{fig:ablation_fd_vs_ad_icemelting} shows that the FD-trained model produces spatially coherent solutions with clean, well-defined interfaces, but the ice phase fraction trajectories exhibit a systematic lag, with the interface evolving consistently more slowly than the reference throughout the simulation. The Allen--Cahn equation contains stiff source terms scaling as $1/\epsilon^2$, and \num{100} time steps over $t\in[0,3]$ cannot accurately resolve the resulting interface dynamics, so the model learns a slightly lagged version of the continuous melting process. AD imposes no such discretization constraint and evaluates residuals exactly at sparse freely sampled collocation points, achieving both higher accuracy and substantially lower training cost. We note that this temporal resolution bottleneck is not exclusive to the FD-based PI-CViT variant studied here. PI-FNO similarly relies on a fixed spatiotemporal grid with finite-difference residuals, and the same limitation may partly account for its consistently higher errors across benchmarks.

\paragraph{Time FiLM conditioning.}
\label{sec:ablation_film}

\begin{figure}[h]
	\centering
	\includegraphics{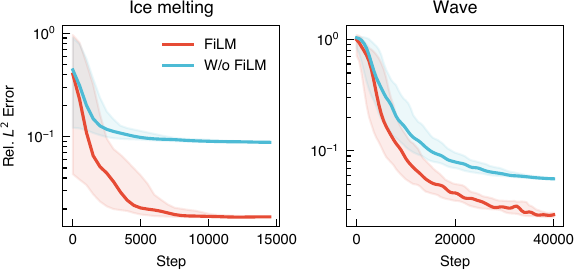}
	\caption{Ablation study on time FiLM conditioning. Test error convergence curves during training for PI-CViT with and without FiLM-modulated decoder on the ice melting and wave benchmarks.}
	\label{fig:ablation_film}
\end{figure}

We evaluate FiLM conditioning on the ice melting and wave benchmarks; results are shown in Figure~\ref{fig:ablation_film}.
FiLM yields clear gains on both, reducing the final error from approximately \qty{9}{\percent} to \qty{2}{\percent} on ice melting and from \qty{6}{\percent} to \qty{3}{\percent} on the wave equation.
In both problems, the spatial pattern of the solution serves as a stable structural template that is globally modulated by time.
For ice melting, the elliptical interface retains its shape throughout the simulation and time governs only the rate of melting.
For the wave equation, the spatial eigenmodes of the heterogeneous medium are determined by the initial displacement and the spatially varying wave speed $c(\mathbf{x})$. The temporal coordinate controls the propagation of wave fronts along these fixed modes without fundamentally altering their spatial structure.
FiLM is well-suited to this class of problems. By using spatial Fourier embeddings as the initial query and injecting temporal information as learned layer-wise scales and shifts, the decoder first establishes the spatial pattern via cross-attention, and temporal information is then applied as a coherent modulation.
For problems with intricate space-time coupling such as Burgers' equation and the shallow water equations, this spatial-temporal decoupling is less appropriate.

\section{Conclusion}
\label{sec:conclusion}

We presented a systematic empirical study of physics-informed neural operator training for parametric PDEs, examining how architecture design, optimizer choice, loss weighting, collocation sampling, derivative evaluation, and time conditioning affect training stability and predictive accuracy across three representative operator backbones and five benchmark problems.

Among the three architectures, PI-CViT consistently achieves the lowest relative $L^2$ errors. On Burgers' equation, PI-CViT (\qty{0.78}{\percent}) improves over PI-FNO (\qty{9.67}{\percent}) by more than an order of magnitude and over PI-DeepONet (\qty{34.5}{\percent}) by a factor of over \num{40}. The gap widens further on the ice melting problem, where both PI-FNO (\qty{52.9}{\percent}) and PI-DeepONet (\qty{37.1}{\percent}) struggle to capture the phase-field dynamics, while PI-CViT remains accurate at \qty{1.87}{\percent}. These results suggest that transformer-based input encoding combined with coordinate-based cross-attention decoding is particularly well suited to physics-informed operator learning.
Our results further show that several optimization pathologies previously identified in PINN training arise naturally in the PINO setting. For instance, on the shallow water benchmark, the gradient norm of the momentum residual exceeds that of the initial-velocity loss by more than two orders of magnitude, and the gradient alignment score under Adam collapses to near zero after early training steps. Corresponding strategies developed for PINNs transfer effectively to this operator-learning setting. GradNorm and causal weighting improve accuracy over unweighted baselines, and the SOAP optimizer reducing the error of PI-CViT by over an order of magnitude on Burgers and wave benchmarks compared to Adam.

We also found that the way the physics loss is evaluated is critical. Freely sampled collocation points provide a richer training signal than residuals restricted to labeled trajectory coordinates, because they allow the model to probe the parametric and spatiotemporal domain beyond a finite dataset. As a result, a well-resolved purely physics-informed training pipeline can match, and in some cases outperform, purely data-driven training across a range of labeled-data regimes. In contrast, simply adding supervised data to the physics loss does not always improve performance, suggesting that hybrid data--physics training requires more careful coordination than a straightforward additive objective.

Despite these promising results, several challenges and opportunities remain for future work.  Extending physics-informed operator learning to irregular domains, complex geometries, and unstructured discretizations will require more geometry-aware architectures and sampling strategies. Scaling to fully three-dimensional problems is another important step, where the computational advantages of operator learning may be especially valuable but where memory cost, optimization stiffness, and residual evaluation become more challenging. The limited synergy observed between supervised and physics-informed losses also motivates future work on training strategies that explicitly manage conflicts between these objectives. Finally, for problems with sharp interfaces, stiff source terms, or strong multi-field coupling, standard pointwise residual losses may be insufficient; variational, conservative, energy-based, or pseudo-time-stepping formulations may provide more faithful physical training signals. We believe that exploring these directions will provide new insights into the design and optimization of physics-informed neural operators and pave the way toward more reliable, data-efficient, and broadly applicable operator-learning methods for scientific computing.


\section*{Acknowledgments}
\label{sec:acknowledgments}
This work was supported by the National Natural Science Foundation of China (NSFC) under Grant No. 52478199. Nanxi Chen acknowledges support from the China Association for Science and Technology (CAST) through the Young Science and Technology Talent Cultivation Project (Doctoral Student Program).

\bibliographystyle{unsrt}
\bibliography{bibfile-nanxi,misc,references}

\appendix

\setcounter{figure}{0}
\setcounter{table}{0}

\renewcommand{\thetable}{A\arabic{table}}
\renewcommand{\thefigure}{A\arabic{figure}}
\section{Nomenclature}
\label{app:notation}

\begin{longtblr}[
	caption = {Notation used throughout the paper.},
	label   = {tab:notation},
]{
	width   = \linewidth,
	colspec = {Q[l,1.2]Q[l,1.2]Q[l,4]},
	row{1}  = {font=\bfseries, bg=gray!30},
	rowhead = 1,
}
	\hline
	Notation & Type & Description \\
	\hline
	\SetCell[c=3]{l,font=\bfseries,bg=gray!15} Domain and geometry \\
	\hline
	$\Omega$ & Set & Spatial domain \\
	$\partial\Omega$ & Set & Boundary of the spatial domain \\
	$[0,T]$ & Interval & Temporal domain \\
	$(\mathbf{x}, t)$ & Tuple & Spatiotemporal coordinate \\
	$L_x,\;L_y$ & Scalar & Spatial periods for the periodic boundary embedding \\
	\hline
	\SetCell[c=3]{l,font=\bfseries,bg=gray!15} PDE formulation \\
	\hline
	$\mathcal{U}$ & Function space & Space of input functions (e.g., initial conditions) \\
	$\mathcal{S}$ & Function space & Solution space of the PDE \\
	$u(\cdot)$ & Function & Input function parameterizing a PDE instance \\
	$s(\mathbf{x}, t; u)$ & Function & PDE solution field corresponding to input $u$ \\
	$\mathcal{N}(\cdot\,; u)$ & Operator & Parameterized PDE differential operator for input $u$ \\
    $\mathcal{I}(\cdot\,; u)$ & Operator & Initial-condition operator for input $u$ \\
    $\mathcal{B}(\cdot\,; u)$ & Operator & Boundary-condition operator for input $u$ \\
	\hline
	\SetCell[c=3]{l,font=\bfseries,bg=gray!15} Neural network architecture \\
	\hline
	$\mathcal{G}_{\boldsymbol{\theta}}: \mathcal{U} \to \mathcal{S}$ & Operator & Neural operator approximating the solution operator \\
	$\boldsymbol{\theta}$ & Set & Trainable parameters of the neural operator \\
	$C,\;H,\;W$ & Scalar & Input field number of channels, height, and width \\
	$P_H,\;P_W$ & Scalar & Patch height and width \\
	$N_p$ & Scalar & Total number of patches ($= H/P_H \times W/P_W$) \\
	$D_e$, $D_d$ & Scalar & Encoder/decoder embedding dimension \\
	$L_e$, $L_d$ & Scalar & Number of encoder/decoder layers \\
	$N_q$ & Scalar & Number of query points in the decoder \\
	\hline
	\SetCell[c=3]{l,font=\bfseries,bg=gray!15} Collocation sampling \\
	\hline
	$M$ & Scalar & Number of input functions sampled per training batch \\
	$N_r$, $N_{\mathrm{ic}}$, $N_{\mathrm{bc}}$ & Scalar & Number of collocation points for the PDE residual, initial condition, and boundary condition losses  \\
	$\{(\mathbf{x}_i, t_i)\}_{i=1}^{N_r}$ & Set & PDE residual collocation points \\
	$\{\mathbf{x}_i\}_{i=1}^{N_{\mathrm{ic}}}$ & Set & Initial condition collocation points \\
	$\{(\mathbf{x}_i, t_i)\}_{i=1}^{N_{\mathrm{bc}}}$ & Set & Boundary condition collocation points \\
	\hline
	\SetCell[c=3]{l,font=\bfseries,bg=gray!15} Loss formulation \\
	\hline
	$\mathcal{L}$ & Scalar & Total training loss \\
	$\mathcal{L}_r,\;\mathcal{L}_{\mathrm{ic}},\;\mathcal{L}_{\mathrm{bc}}$ & Scalar & PDE residual, IC, and BC loss components \\
	$w_r,\;w_{\mathrm{ic}},\;w_{\mathrm{bc}}$ & Scalar & Scalar weights for each loss component \\
	$\mathcal{J}_{\mathcal{L}}$ & Set & Index set of all loss terms \\
	$g_\ell^{(s)}$ & Vector & Gradient of loss $\ell$ w.r.t.\ $\boldsymbol{\theta}$ at training step $s$ \\
    $\mathcal{A}^{(s)}$ & Scalar & Gradient alignment score at training step $s$ \\
	$\hat{w}_\ell^{(s)}$, $w_\ell^{(s)}$ & Scalar & Raw and EMA-smoothed GradNorm weight for loss $\ell$ \\
	$\alpha_w$ & Scalar & EMA smoothing factor for loss weights \\
	$N_t$ & Scalar & Number of temporal segments for causal weighting \\
	$w_{\mathrm{causal}}^{(i)}$ & Scalar & Causal weight for the $i$-th time segment \\
	$\epsilon$ & Scalar & Causal strength hyperparameter \\
	\hline
	\SetCell[c=3]{l,font=\bfseries,bg=gray!15} Training and Optimization \\
	\hline
	$\eta$ & Scalar & Learning rate \\
	$G^{(s)}$ & Matrix & Raw gradient matrix at training step $s$ \\
	$Q_L,\;Q_R$ & Matrix & Left/right eigenvector matrices of the gradient second-moment \\
	$\Delta\widetilde{W}$ & Matrix & Parameter update in the preconditioned eigenbasis (SOAP) \\
	\hline
\end{longtblr}

\section{Evaluation metric}
\label{app:metric}

We employ the relative $L^2$ error as the primary evaluation metric throughout all benchmarks. The error is computed on a discrete set of evaluation points and averaged over channels and test samples. For the time-dependent problems, the reference test trajectories are stored at $\Delta t = 0.1$. PI-CViT and PI-DeepONet are evaluated directly on this spatiotemporal mesh, whereas PI-FNO first predicts on its native grid with $\Delta t = T/100$ and is then temporally subsampled at the stored reference times before error computation:
\begin{equation}
	\text{Rel. } L^2 = \frac{1}{N_{\mathrm{test}}} \sum_{i=1}^{N_{\mathrm{test}}} \frac{1}{C} \sum_{c=1}^{C} \frac{\left\lVert\mathcal{G}_{\boldsymbol{\theta}}(u_i)^{(c)} - s^{(c)}(\cdot\,; u_i)\right\rVert_{\Omega\times[0,T]}}{\left\lVert s^{(c)}(\cdot\,; u_i)\right\rVert_{\Omega\times[0,T]}},
\end{equation}
where $C$ is the number of solution channels, $N_{\mathrm{test}} = 100$, and
\begin{equation}
\left\lVert s^{(c)} \right\rVert_{\Omega\times[0,T]} = \sqrt{\sum_{k=1}^{N_x}\sum_{l=1}^{N_t} \bigl|s^{(c)}(\mathbf{x}_k, t_l)\bigr|^2}.
\end{equation}
Here $\{\mathbf{x}_k\}_{k=1}^{N_x}$ are all mesh points of the reference numerical solver, $\{t_l\}_{l=1}^{N_t}$ are temporal snapshots at uniform intervals of $\Delta t = 0.1$, and the evaluation is performed at all Cartesian product points $\{(\mathbf{x}_k, t_l)\}$, yielding $N_x \times N_t$ spatiotemporal evaluation points in total. For steady-state lid-driven cavity flow, the temporal dimension is omitted.



\section{Parametric input families}
\label{app:parametric_inputs}
This section summarizes the parametric input distributions used to define the benchmark PDE families. The considered problems span three representative types of inputs: smooth random fields for the Burgers, wave, and shallow water equations; structured geometric parameters for the ice melting problem; and a single scalar physical parameter for the lid-driven cavity flow. These choices are intended to test physics-informed neural operators across different levels of input complexity, ranging from high-dimensional function-to-function mappings to low-dimensional parameter-to-field mappings.

\subsection{Periodic Gaussian random fields}
\label{app:grf}

We generate initial conditions for the Burgers, wave, and shallow water equations by sampling from a zero-mean \emph{periodic} Gaussian random field (GRF) with a Gaussian power spectrum, implemented via a spectral method. For a spatial grid of resolution $N_x \times N_y$ on a periodic domain $[0, L_x] \times [0, L_y]$, the discrete angular wavenumbers are
\begin{equation}
	k_{x, n_x} = \frac{2\pi n_x}{L_x}, \qquad k_{y, n_y} = \frac{2\pi n_y}{L_y},
\end{equation}
where $n_x \in \{0, 1, \ldots, \lfloor N_x/2\rfloor-1, -\lfloor N_x/2\rfloor, \ldots, -1\}$ and analogously for $n_y$. The squared wavenumber magnitude at each mode is $|\mathbf{k}|^2 = k_x^2 + k_y^2$.

A Gaussian power spectrum is defined in wavenumber space as
\begin{equation}
	S(|\mathbf{k}|^2) = \exp\!\left(-\frac{\ell^2\,|\mathbf{k}|^2}{2}\right),
\end{equation}
where $\ell$ is the length-scale parameter controlling spatial smoothness: larger $\ell$ yields smoother, more correlated fields, while smaller $\ell$ produces rougher fields. Complex Fourier coefficients are then drawn as
\begin{equation}
	\hat{\xi}(\mathbf{k}) = \bigl(\xi_r(\mathbf{k}) + \mathrm{i}\,\xi_i(\mathbf{k})\bigr)\sqrt{S(|\mathbf{k}|^2)},
\end{equation}
where $\xi_r(\mathbf{k}),\,\xi_i(\mathbf{k}) \stackrel{\text{i.i.d.}}{\sim} \mathcal{N}(0,1)$ are independent standard Gaussian random variables. The physical-space field is recovered via the inverse discrete Fourier transform,
\begin{equation}
	\xi(\mathbf{x}) = \mathrm{Re}\!\left[\mathcal{F}^{-1}\!\left\{\hat{\xi}\right\}(\mathbf{x})\right], \label{eq:grf}
\end{equation}
and is standardized to zero mean and unit variance before being scaled by a prescribed amplitude $A$:
\begin{equation}
    \xi\;\leftarrow\; A \cdot \frac{\xi - \mu_\xi}{\sigma_\xi},
\end{equation}
where $\mu_\xi$ and $\sigma_\xi$ denote the sample mean and standard deviation. 

For all three benchmarks parameterized by GRF initial conditions, the length scale is fixed at $\ell=0.1$. The amplitude $A$ is set to \num{0.2}, \num{0.5}, and \num{0.2} for the Burgers', wave, and shallow water equations, respectively, to keep the solution fields within a physically reasonable range without requiring additional normalization or rescaling.

\subsection{Rotated elliptical inclusions}
\label{app:ellipse}

The parametric family for the ice melting problem consists of rotated elliptical inclusions. Each instance is described by semi-axes $a, b > 0$ and a rotation angle $\theta \in [0, \pi)$. For a point $\mathbf{x} = (x, y)$, the coordinates aligned with the ellipse axes are obtained by rotating by $-\theta$:
\begin{equation}
	\begin{pmatrix} x' \\ y' \end{pmatrix}
	=
	\begin{pmatrix} \cos\theta & \sin\theta \\ -\sin\theta & \cos\theta \end{pmatrix}
	\begin{pmatrix} x \\ y \end{pmatrix}.
\end{equation}
An approximate signed distance from $\mathbf{x}$ to the ellipse boundary is then defined as
\begin{equation}
	d(\mathbf{x}) \approx \frac{2ab}{a+b}\left[1 - \sqrt{\left(\frac{x'}{a}\right)^2 + \left(\frac{y'}{b}\right)^2}\,\right],
	\label{eq:ellipse_dist}
\end{equation}
so that $d > 0$ inside the ellipse and $d < 0$ outside. The prefactor $2ab/(a+b)$ is the harmonic mean of the semi-axes; for a circle ($a = b = r$) the formula reduces to the exact signed distance $r - \|\mathbf{x}\|$, and for a general ellipse it provides a first-order accurate approximation in the vicinity of the boundary. All ellipses are centered at the origin, and the parameters are uniformly sampled as
\begin{align}
	a,\, b &\sim \mathcal{U}(20,40), \\
	\theta &\sim \mathcal{U}(0, \pi).
\end{align}

The diffuse-interface initial condition is then set as
\begin{equation}
	\phi(\mathbf{x}, 0) = \tanh\!\left(\frac{d(\mathbf{x})}{\sqrt{2}\,\epsilon}\right),
\end{equation}
which transitions smoothly from $\phi \approx +1$ (inside the ellipse) to $\phi \approx -1$ (outside) over an interface of width $O(\epsilon)$.

\subsection{Log-uniform Reynolds number sampling}
\label{app:re}

The lid-driven cavity problem is parameterized by a single scalar, the Reynolds number $\mathrm{Re}$. Since the solution behavior varies more smoothly on a logarithmic scale than a linear one, we draw samples log-uniformly over the interval $[\mathrm{Re}_{\min}, \mathrm{Re}_{\max}] = [50, 1000]$ using Latin hypercube sampling (LHS):
\begin{equation}
	\log \mathrm{Re} \;\overset{\text{LHS}}{\sim}\; \mathcal{U}\!\left(\log \mathrm{Re}_{\min},\, \log \mathrm{Re}_{\max}\right).
\end{equation}

Before being passed to the model, each sampled value is normalized to $[0, 1]$ via min-max scaling in the log space:
\begin{equation}
	\widetilde{\mathrm{Re}} = \frac{\log \mathrm{Re} - \log \mathrm{Re}_{\min}}{\log \mathrm{Re}_{\max} - \log \mathrm{Re}_{\min}},
\end{equation}
so that $\widetilde{\mathrm{Re}} = 0$ corresponds to $\mathrm{Re} = 50$ and $\widetilde{\mathrm{Re}} = 1$ corresponds to $\mathrm{Re} = 1000$. This normalized scalar serves as the input encoding for the operator.

\section{Numerical solvers for reference solutions}
\label{app:numerical_solvers}

Reference solutions for all benchmark PDEs are computed using dedicated numerical solvers: a pseudo-spectral method for the Burgers', wave, and shallow water equations; a finite-element method for the ice melting problem; and a finite-difference method for the lid-driven cavity flow. We generate 100 test solutions for each benchmark with randomly sampled input functions as described in Appendix~\ref{app:parametric_inputs}.

\paragraph{Burgers' equation.}
\label{app:burgers_solver}

Reference solutions are generated using a pseudo-spectral method with a classical fourth-order Runge--Kutta (RK4) time integrator. The velocity field $\mathbf{v} = (v_1, v_2)$ is represented on a uniform $N_x \times N_y$ grid over the doubly periodic domain $[0,L_x)\times[0,L_y)$. Spatial derivatives are evaluated spectrally: for any field $w$ with 2D discrete Fourier transform $\hat{w}$,
\begin{equation}
	\widehat{\partial_x w}(\mathbf{k}) = ik_x\,\hat{w}(\mathbf{k}), \qquad
	\widehat{\partial_y w}(\mathbf{k}) = ik_y\,\hat{w}(\mathbf{k}),
\end{equation}
where $k_x = 2\pi n_x/L_x$ and $k_y = 2\pi n_y/L_y$. The right-hand side combines nonlinear advection and viscous diffusion:
\begin{equation}
	\widehat{\mathcal{R}[\mathbf{v}]}(\mathbf{k}) = -\widehat{(\mathbf{v}\cdot\nabla)\mathbf{v}}(\mathbf{k}) - \nu|\mathbf{k}|^2\,\hat{\mathbf{v}}(\mathbf{k}).
\end{equation}
The nonlinear term is evaluated by pointwise multiplication in physical space and then transformed to Fourier space. To suppress aliasing errors from the quadratic interactions, the 2/3 truncation rule is applied: all modes with $|k_x| > \tfrac{2}{3}k_{x,\max}$ or $|k_y| > \tfrac{2}{3}k_{y,\max}$ are zeroed. The solution is advanced by
\begin{equation}
	\mathbf{v}^{n+1} = \mathbf{v}^n + \frac{\Delta t}{6}\bigl(\kappa_1 + 2\kappa_2 + 2\kappa_3 + \kappa_4\bigr),
\end{equation}
where $\kappa_1 = \mathcal{R}(\mathbf{v}^n)$, $\kappa_2 = \mathcal{R}(\mathbf{v}^n + \frac{\Delta t}{2}\kappa_1)$, $\kappa_3 = \mathcal{R}(\mathbf{v}^n + \frac{\Delta t}{2}\kappa_2)$, $\kappa_4 = \mathcal{R}(\mathbf{v}^n + \Delta t\,\kappa_3)$. Simulations use $N_x = N_y = 64$, $L_x = L_y = 1.0$, $\nu = 0.01$, $\Delta t = \num{1e-3}$, and $t\in[0,1]$.

\paragraph{Wave equation.}

The spatial discretization and RK4 time integration follow the same pseudo-spectral framework as in Appendix~\ref{app:burgers_solver}. The second-order wave equation is recast as a first-order system by introducing $\psi = \partial_t v$, giving state vector $\mathbf{q} = (v, \psi)$ and right-hand side
\begin{equation}
	\mathcal{R}[v,\psi] = \bigl(\psi,\;\; c(\mathbf{x})^2\nabla^2 v\bigr),
\end{equation}
where the Laplacian is evaluated spectrally via $\widehat{\nabla^2 v}(\mathbf{k}) = -|\mathbf{k}|^2\hat{v}(\mathbf{k})$, and the spatially varying factor $c(\mathbf{x})^2$ is applied as a pointwise multiplication in physical space. Since the equation is linear in $v$, no dealiasing is required. Simulations use $N_x = N_y = 64$, $L_x = L_y = 1$, $\Delta t = \num{5e-4}$, and $t\in[0,1]$.

\paragraph{Shallow water equation.}

The spatial discretization and RK4 time integration follow the same pseudo-spectral framework as in Appendix~\ref{app:burgers_solver}. The state vector is $\mathbf{q} = (h, v_1, v_2)$. Since the linearized shallow water equations are entirely linear, all spatial derivatives are computed spectrally without dealiasing, and the right-hand side reads directly in Fourier space:
\begin{align}
	\widehat{\mathcal{R}[h]}(\mathbf{k}) &= -H\bigl(ik_x\hat{v}_1 + ik_y\hat{v}_2\bigr), \\
	\widehat{\mathcal{R}[v_1]}(\mathbf{k}) &= f\hat{v}_2 - g\,ik_x\hat{h}, \\
	\widehat{\mathcal{R}[v_2]}(\mathbf{k}) &= -f\hat{v}_1 - g\,ik_y\hat{h},
\end{align}
where $H$ is the mean fluid depth, $g$ the gravitational acceleration, and $f$ the Coriolis parameter. Simulations use $N_x = N_y = 64$, $L_x = L_y = 1$, $H = g =1$,  $f = 10$, $\Delta t = \num{1e-3}$, and $t\in[0,1]$.

\paragraph{Ice melting.}

The Allen--Cahn equation is discretized using the finite element method (FEM) with a fully implicit (backward Euler) time-stepping scheme, implemented in \texttt{FEniCS}. The spatial domain $\Omega=[-50, 50]^2$ is partitioned into a uniform $N\times N$ rectangular mesh using continuous piecewise-linear (P1/CG1) elements, giving the finite element space $V_h \subset H^1(\Omega)$.

Multiplying the governing equation by a test function $v \in V_h$ and integrating over $\Omega$ yields the weak form. The Laplacian term is integrated by parts; the boundary term vanishes due to the homogeneous Neumann condition $\nabla\phi\cdot\mathbf{n}|_{\partial\Omega} = 0$. The semi-discrete weak form reads: find $\phi\in V_h$ such that for all $v\in V_h$,
\begin{equation}
	\int_\Omega \frac{\partial\phi}{\partial t}\,v\,\mathrm{d}\mathbf{x}
	+ M\int_\Omega \nabla\phi\cdot\nabla v\,\mathrm{d}\mathbf{x}
	+ \frac{M}{\epsilon^2}\int_\Omega F'(\phi)\,v\,\mathrm{d}\mathbf{x}
	+ \frac{\lambda}{\epsilon}\int_\Omega \sqrt{2F(\phi)}\,v\,\mathrm{d}\mathbf{x} = 0,
\end{equation}
where $F(\phi) = \tfrac{1}{4}(\phi^2-1)^2$ and $F'(\phi) = \phi^3-\phi$. Applying a backward Euler approximation to the time derivative gives the fully discrete problem at each time step: find $\phi^{n+1}\in V_h$ such that for all $v\in V_h$,
\begin{equation}
	\int_\Omega \frac{\phi^{n+1}-\phi^n}{\Delta t}\,v\,\mathrm{d}\mathbf{x}
	+ M\!\int_\Omega \nabla\phi^{n+1}\cdot\nabla v\,\mathrm{d}\mathbf{x}
	+ \frac{M}{\epsilon^2}\!\int_\Omega F'(\phi^{n+1})\,v\,\mathrm{d}\mathbf{x}
	+ \frac{\lambda}{\epsilon}\!\int_\Omega \sqrt{2F(\phi^{n+1})}\,v\,\mathrm{d}\mathbf{x} = 0.
	\label{eq:ac_fem_discrete}
\end{equation}
The nonlinear system~\eqref{eq:ac_fem_discrete} at each time step is solved by Newton's method with the \texttt{MUMPS} direct solver. Simulations use $N = 63$ (yielding a $64\times 64$ uniform grid), $M = 0.1$, $\lambda = 5$, interface thickness $\epsilon = 3h/(\sqrt{2}\,\mathrm{arctanh}(0.9))$ with mesh size $h = 100/N$, time step $\Delta t=\num{1e-3}$, and $t\in[0,3]$.

\paragraph{Lid-driven cavity flow.}

The steady lid-driven cavity flow is computed via the vorticity--streamfunction formulation~\cite{gangwar2026solver}. Introducing the streamfunction $\psi$ such that $v_1 = \partial\psi/\partial y$ and $v_2 = -\partial\psi/\partial x$, and the vorticity $\omega = \partial v_2/\partial x - \partial v_1/\partial y = -\nabla^2\psi$, the incompressible Navier--Stokes equations reduce to two coupled scalar problems:
\begin{align}
	\nabla^2\psi &= -\omega, \label{eq:poisson_psi} \\
	\frac{\partial\omega}{\partial t} + v_1\frac{\partial\omega}{\partial x} + v_2\frac{\partial\omega}{\partial y} &= \nu\nabla^2\omega. \label{eq:vorticity_transport}
\end{align}
The domain $[0,1]^2$ is discretized on a uniform $N\times N$ grid with spacing $h = 1/(N-1)$. Velocities are recovered from $\psi$ by second-order central differences. Convective terms in~\eqref{eq:vorticity_transport} are discretized by first-order upwind differences and diffusive terms by second-order central differences. The steady solution is sought by iterating the following two-stage procedure until the vorticity change satisfies $\max|\Delta\omega| < 10^{-6}$. At each iteration, the vorticity field is first advanced by one time step $\Delta t = 0.1h/U$ using the Peaceman--Rachford alternating direction implicit (ADI) scheme~\cite{peaceman1955numerical}, with tridiagonal sub-systems solved via the Thomas algorithm. The updated vorticity is then used as the source in the Poisson equation~\eqref{eq:poisson_psi}, which is solved with red-black successive over-relaxation (SOR) (relaxation factor $\omega_{\mathrm{SOR}} = 1.8$) until $\max|\Delta\psi| < 10^{-4}$.

The streamfunction satisfies $\psi = 0$ on all four walls. Vorticity values at solid boundaries are updated each outer iteration using Thom's formula~\cite{thom1933flow}:
\begin{gather}
	\omega_{N,j} = -\frac{2\psi_{N-1,j}}{h^2} - \frac{2U}{h} \quad \text{(top moving lid)}, \\
	\omega_{1,j} = -\frac{2\psi_{2,j}}{h^2}, \quad
	\omega_{i,1} = -\frac{2\psi_{i,2}}{h^2}, \quad
	\omega_{i,N} = -\frac{2\psi_{i,N-1}}{h^2} \quad \text{(stationary walls)},
\end{gather}
where $U$ is the lid velocity. Corner nodes are assigned the average of their two adjacent wall values. To capture the sharp flow features at high Reynolds numbers, simulations use $N = 256$, $U=1$, and $\nu  = 1/\mathrm{Re}$ with parametric Reynolds numbers sampled as described in Appendix~\ref{app:re}.

\section{Hyperparameters}
\label{app:hyperparameters}

\subsection{Training hyperparameters}
\label{app:training_hyperparameters}
The training hyperparameters used for all experiments and shared across all three architectures are summarized in Table~\ref{tab:training_hyperparameters}. 
All benchmarks use an exponential decay learning rate schedule; the lid-driven cavity is the exception, using a schedule-free training scheme~\cite{defazio2024road} instead. SOAP is the default optimizer for all benchmarks, with Adam substituted for the ice melting problem for better stability. Causal weighting is applied to all time-dependent benchmarks and omitted for the steady-state lid-driven cavity flow. We resample the input functions and collocation points at every training step; during the warm-up phase, the resampling frequency is reduced to every \num{50}--\num{100} steps to stabilize convergence of the initial condition loss. 
The coordinate sampling parameters apply only to PI-CViT and PI-DeepONet, as PI-FNO operates on a fixed discretization grid and does not use collocation points. For the initial condition, we sample a dense grid of $64^2$ points matching the input field resolution to fully capture the initial state. For problems with additional initial conditions (e.g., the velocity components in the wave and shallow water equations), we use a smaller number of collocation points for those terms. Periodic boundary conditions are enforced via the periodic embedding in \eqref{eq:periodic_embedding} and do not require explicit sampling or loss terms; for the ice melting problem, the entire boundary remains in the liquid phase with spatially uniform field values, so the homogeneous Neumann boundary conditions are trivially satisfied and require no explicit enforcement; for the lid-driven cavity flow, we sample \num{256} points on each of the four walls to enforce the velocity and no-slip boundary conditions.
\begin{table}[htbp]
	\centering
	\caption{Training hyperparameters for all benchmarks. SW = shallow water, IM = ice melting, LDC = lid-driven cavity.}
	\label{tab:training_hyperparameters}
	\begin{tblr}{
		width = \textwidth,
		colspec = {Q[l,2.5]Q[c,1]Q[c,1]Q[c,1]Q[c,1]Q[c,1]},
		row{1} = {font=\bfseries,bg=gray!30},
		}
		\hline
		Parameter                        & Burgers     & Wave        & SW          & IM          & LDC           \\
		\hline
		\SetCell[c=6,r=1]{l,font=\bfseries,bg=gray!15}Learning rate schedule                                     \\
		\hline
		Initial learning rate            & \num{5e-4}  & \num{5e-4}  & \num{5e-4}  & \num{5e-4}  & \num{5e-4}    \\
		Decay rate                       & \num{0.95}  & \num{0.95}  & \num{0.95}  & \num{0.95}  & N/A    \\
		Decay steps                      & \num{200}   & \num{500}   & \num{500}   & \num{200}   & N/A     \\	
		Minimum learning rate            & \num{1e-5}  & \num{1e-5}  & \num{1e-6}  & \num{1e-5}  & \num{1e-5}    \\
		\hline
		\SetCell[c=6,r=1]{l,font=\bfseries,bg=gray!15}Optimizer settings                                     \\
		\hline
		Optimizer name                      & SOAP        & SOAP        & SOAP        & Adam        & SOAP          \\
		Precondition frequency (SOAP)            &  \num{5} \ & \num{5}     & \num{5}     & N/A         & \num{5}       \\
		EMA factor $b_1$ & \num{0.95} & \num{0.95}  & \num{0.90}  & \num{0.9}   & \num{0.95}    \\
		EMA factor $b_2$ & \num{0.95}  & \num{0.95}   & \num{0.90}   & \num{0.999} & \num{0.95}     \\
		Weight Decay & \num{1e-6} & \num{1e-6} & \num{0} & \num{1e-3} & \num{1e-6} \\
		\hline
		\SetCell[c=6,r=1]{l,font=\bfseries,bg=gray!15}Training and sampling                                      \\
		\hline
		Total training steps             & \num{20000} & \num{40000} & \num{25000} & \num{15000} & \num{15000}   \\
		Warm-up steps                    & \num{1500}  & \num{1500}  & \num{1000}  & \num{500}  & \num{0}           \\
		Num. of input functions & \num{32}    & \num{32}    & \num{32}    & \num{16}    & \num{32}           \\
		Num. of residual samplings       & \num{2048}  & \num{2048}  & \num{2048}  & \num{4096}  & \num{2048}           \\
		Num. of IC samplings (for $u$)   & $64^2$      & $64^2$      & $64^2$      & $64^2$      & N/A           \\
		Num. of IC samplings (others)    & N/A         & \num{1024}  & \num{2048}  & N/A         & N/A           \\
		Num. of BC samplings             & N/A         & N/A         & N/A         & N/A         & \numproduct{256x4} \\
		\hline
		\SetCell[c=6,r=1]{l,font=\bfseries,bg=gray!15}GradNorm and causal weighting                                   \\
		\hline
		GradNorm EMA factor            & \num{1.0}    & \num{1.0}   & \num{1.0}    & \num{1.0}    & \num{1.0}         \\
		Num. of causal time segments            & \num{24}    & \num{24}    & \num{24}    & \num{24}    & N/A           \\
		Initial causal strength          & \num{1e-2}   & \num{1e-2}  & \num{1e-5}   & \num{1e-2}   & N/A           \\
		Maximum causal strength          & \num{5.0}   & \num{10.0}  & \num{1.0}   & \num{10.0}   & N/A           \\
		\hline
	\end{tblr}
\end{table}

\subsection{Architectural hyperparameters}
\label{app:architectural_hyperparameters}

\paragraph{PI-CViT.}
Table~\ref{tab:architectural_hyperparameters_cvit} summarizes the architectural hyperparameters for PI-CViT across all benchmarks. For problems parameterized by spatial fields, the encoder architecture is largely consistent, while for lid-driven cavity flow, which is parameterized by a scalar Reynolds number, the encoder is simplified to a small MLP without attention layers. 
\begin{table}[h]
	\centering
	\caption{Architectural hyperparameters for PI-CViT on all benchmarks. SW = shallow water, IM = ice melting, LDC = lid-driven cavity.}
	\label{tab:architectural_hyperparameters_cvit}
	\begin{tblr}{
		width = \textwidth,
		colspec = {Q[l,2.5]Q[c,1]Q[c,1]Q[c,1]Q[c,1]Q[c,1]},
		row{1} = {font=\bfseries,bg=gray!30},
		}
		\hline
		Parameter                    & Burgers            & Wave               & SW                 & IM                 & LDC          \\
		\hline
		\SetCell[c=6,r=1]{l,font=\bfseries,bg=gray!15}Encoder                                                                           \\
		\hline
		Grid size                    & \numproduct{64x64} & \numproduct{64x64} & \numproduct{64x64} & \numproduct{64x64} & N/A          \\
		Patch size                   & \numproduct{8x8}   & \numproduct{8x8}   & \numproduct{8x8}   & \numproduct{8x8}   & N/A          \\
		Encoder depth                & \num{2}            & \num{2}            & \num{2}            & \num{2}            & \num{2}      \\
		Encoder dim.                 & \num{256}          & \num{256}          & \num{256}          & \num{256}          & \num{256}    \\
		Encoder num. of heads        & \num{8}            & \num{8}            & \num{8}            & \num{8}            & N/A          \\
		\hline
		\SetCell[c=6,r=1]{l,font=\bfseries,bg=gray!15}Decoder                                                                           \\
		\hline
		Fourier features freq. for $\mathbf{x}$       & \num{2.0}          & \num{2.0}          & \num{2.0}          & \num{1.0}          & \num{2.0}    \\
		Fourier features freq. for $t$      & \num{2.0}          & \num{2.0}          & \num{2.0}          & \num{1.0}          & \num{2.0}    \\
		Decoder depth                & \num{2}            & \num{4}            & \num{4}            & \num{2}            & \num{4}      \\
		Decoder dim.                 & \num{256}          & \num{384}          & \num{384}          & \num{256}          & \num{256}    \\
		Decoder num. of heads        & \num{8}            & \num{12}           & \num{8}            & \num{8}            & \num{8}      \\
		Num. MLP layers              & \num{2}            & \num{2}            & \num{2}            & \num{3}            & \num{2}      \\
		Activation                   & GELU               & GELU               & GELU               & Tanh               & GELU         \\
		Time FiLM                    & False              & True               & False              & True               & N/A          \\
		\hline
		\SetCell[c=6,r=1]{l,font=\bfseries,bg=gray!15}Others                                                                            \\
        \hline
		Num. of trainable parameters & \qty{3.4}{M}       & \qty{10.6}{M}      & \qty{9.0}{M}       & \qty{3.9}{M}       & \qty{3.5}{M} \\
		\hline
	\end{tblr}
\end{table}

\paragraph{PI-DeepONet}
PI-DeepONet consists of a branch network and a trunk network, both implemented as fully connected networks with GELU or Tanh activations; architectural details are given in Table~\ref{tab:architectural_hyperparameters_deeponet}. The branch network encodes the flattened initial condition field at the full $64^2$ grid resolution into a feature vector of a specified dimension (i.e., basis dimension). The trunk network maps spatiotemporal query coordinates, augmented by Fourier features embeddings, to a feature vector of the same dimension. The final prediction is obtained as a point-wise inner product of these two vectors. For the lid-driven cavity, the branch network degenerates to a linear layer with an input dimension of 1 to take in the scalar Reynolds number. Network depth and width are tuned across benchmarks to yield a parameter count in the range of 3--10 million, for a fair comparison with PI-CViT.
\begin{table}[h]
	\centering
	\caption{Architectural hyperparameters for PI-DeepONet on all benchmarks. SW = shallow water, IM = ice melting, LDC = lid-driven cavity.}
	\label{tab:architectural_hyperparameters_deeponet}
	\begin{tblr}{
		width = \textwidth,
		colspec = {Q[l,2.5]Q[c,1]Q[c,1]Q[c,1]Q[c,1]Q[c,1]},
		row{1} = {font=\bfseries,bg=gray!30},
		}
		\hline
		Parameter                    & Burgers            & Wave               & SW                 & IM                 & LDC          \\
		\hline
		\SetCell[c=6,r=1]{l,font=\bfseries,bg=gray!15}Branch network                                                                    \\
		\hline
		Branch depth                 & \num{8}            & \num{8}            & \num{6}            & \num{6}            & \num{2}      \\
		Branch dim.                  & \num{256}          & \num{512}          & \num{512}          & \num{256}          & \num{512}    \\
		\hline
		\SetCell[c=6,r=1]{l,font=\bfseries,bg=gray!15}Trunk network                                                                     \\
		\hline
		Fourier features freq. for $\mathbf{x}$       & \num{2.0}          & \num{2.0}          & \num{2.0}          & \num{1.0}          & \num{2.0}    \\
		Fourier features freq. for $t$      & \num{2.0}          & \num{2.0}          & \num{2.0}          & \num{1.0}          & \num{2.0}    \\
		Trunk depth                  & \num{8}            & \num{8}            & \num{6}            & \num{6}            & \num{8}      \\
		Trunk dim.                   & \num{256}          & \num{512}          & \num{512}          & \num{256}          & \num{512}    \\
		Activation                   & GELU               & GELU               & GELU               & Tanh               & Tanh         \\
		\hline
		\SetCell[c=6,r=1]{l,font=\bfseries,bg=gray!15}Others                                                                            \\
        \hline
		Basis dim.                   & \num{256}          & \num{512}          & \num{512}          & \num{256}          & \num{512}    \\
		Num. of trainable parameters & \qty{3.5}{M}       & \qty{7.4}{M}       & \qty{6.0}{M}       & \qty{2.1}{M}       & \qty{3.4}{M} \\
		\hline
	\end{tblr}
\end{table}

\paragraph{PI-FNO.}
PI-FNO processes the input through a point-wise lifting convolution, a stack of Fourier spectral convolution layers each supplemented by a residual point-wise convolution in physical space, and a final projection layer; architectural hyperparameters are given in Table~\ref{tab:architectural_hyperparameters_fno}. For time-dependent benchmarks, convolutions span the full spatiotemporal domain, with grid coordinates concatenated as additional input channels; the entire solution trajectory is predicted in a single forward pass on a fixed spatiotemporal grid with 100 uniform time steps, i.e., $\Delta t = T/100$. At test time, the reported errors are computed on the stored reference snapshots with $\Delta t = 0.1$, obtained by temporally subsampling the PI-FNO prediction.
For the lid-driven cavity, the spectral convolutions are restricted to the spatial domain and the Reynolds number is broadcast as a constant input channel across the spatial grid. Padding is applied along non-periodic axes to suppress spectral wrap-around artifacts: e.g., for lid-driven cavity flow, the spatial domain is padded by 32 grid points on each side; while for time-dependent benchmarks, the temporal domain is padded by 10 time steps at the end~\cite{rosofskyApplicationsPhysicsInformed2023}. Because PI-FNO operates on a fixed volumetric discretization, its parameter count is substantially larger than those of the other two architectures even at a modest number of retained Fourier modes. 
\begin{table}[h]
	\centering
	\caption{Architectural hyperparameters for PI-FNO on all benchmarks. SW = shallow water, IM = ice melting, LDC = lid-driven cavity.}
	\label{tab:architectural_hyperparameters_fno}
	\begin{tblr}{
		width = \textwidth,
		colspec = {Q[l,2.5]Q[c,1]Q[c,1]Q[c,1]Q[c,1]Q[c,1]},
		row{1} = {font=\bfseries,bg=gray!30},
		}
		\hline
		Parameter                     & Burgers       & Wave          & SW            & IM            & LDC           \\
		\hline
		\SetCell[c=6,r=1]{l,font=\bfseries,bg=gray!15}FNO                                                             \\
		\hline
		Num. of Fourier modes for $x$ & \num{8}       & \num{8}       & \num{8}       & \num{8}       & \num{8}       \\
		Num. of Fourier modes for $y$ & \num{8}       & \num{8}       & \num{8}       & \num{8}       & \num{8}       \\
		Num. of Fourier modes for $t$ & \num{8}       & \num{8}       & \num{8}       & \num{8}       & N/A           \\
		Width                         & \num{32}      & \num{64}      & \num{64}      & \num{32}      & \num{64}      \\
		Depth                         & \num{8}       & \num{4}       & \num{4}       & \num{4}       & \num{4}       \\
		Time steps                    & \num{100}     & \num{100}     & \num{100}     & \num{100}     & N/A           \\
		Padding for $(x, y, t)$       & (0, 0, 10)    & (0, 0, 10)    & (0, 0, 10)    & (0, 0, 10)    & (32, 32, N/A) \\
		Activation                    & GELU          & GELU          & GELU          & GELU          & GELU          \\
		\hline
		\SetCell[c=6,r=1]{l,font=\bfseries,bg=gray!15}Others                                                          \\
        \hline
		Num. of trainable parameters  & \qty{33.6}{M} & \qty{67.1}{M} & \qty{67.1}{M} & \qty{16.8}{M} & \qty{4.2}{M}  \\
		\hline
	\end{tblr}
\end{table}

\section{Computational cost}
\label{app:computational_cost}

All models are trained on a machine equipped with two NVIDIA RTX PRO 6000 GPUs (96\,GB memory each) and a 22-core Intel Xeon Platinum 8470Q CPU. All code is implemented in \texttt{JAX} and \texttt{equinox}. The wall-clock training time for each benchmark is summarized in Table~\ref{tab:training_time}. PI-DeepONet is the fastest model owing to its purely linear branch and trunk network architecture, which incurs minimal computational overhead per training step. PI-FNO, by contrast, is considerably slower due to its substantially larger parameter count. Despite its moderate training cost, PI-CViT achieves the highest accuracy across benchmarks (Table~\ref{tab:main_results}).
\begin{table}[h]
	\centering
	\caption{Wall-clock training time in hours per \num{10000} training steps for all benchmarks.
	SW = shallow water, IM = ice melting, LDC = lid-driven cavity.}
	\label{tab:training_time}
	\begin{tblr}{
		width = \textwidth,
		colspec = {Q[c,1.5]Q[c,1]Q[c,1]Q[c,1]Q[c,1]Q[c,1]},
		row{1} = {font=\bfseries},
		}
		\hline
		Models      & Burgers    & Wave        & SW          & IM         & LDC        \\
		\hline
		PI-CViT     & \num{1.0}  & \num{2.41}  & \num{1.6}   & \num{0.65} & \num{0.88} \\
		PI-DeepONet & \num{0.46} & \num{0.55}  & \num{0.27}  & \num{0.1} & \num{0.73}  \\
		PI-FNO      & \num{3.29} & \num{4.23} & \num{4.26} & \num{1.46} & \num{2.20}  \\
		\hline
	\end{tblr}
\end{table}

\section{Additional visualizations}
\label{app:additional_visualizations}

This section provides supplementary results for PI-CViT trained with a pure physics loss. For every benchmark, we include solution trajectories for additional test samples beyond the single representative sample shown in the main text. For selected benchmarks, we further provide complementary views such as derived fields computed from the predicted solutions (e.g., vorticity and divergence) and more detailed visualizations such as spatial slices that reveal fine-grained spatiotemporal structure.

\begin{figure}[h]
	\centering
	\includegraphics[width=\textwidth]{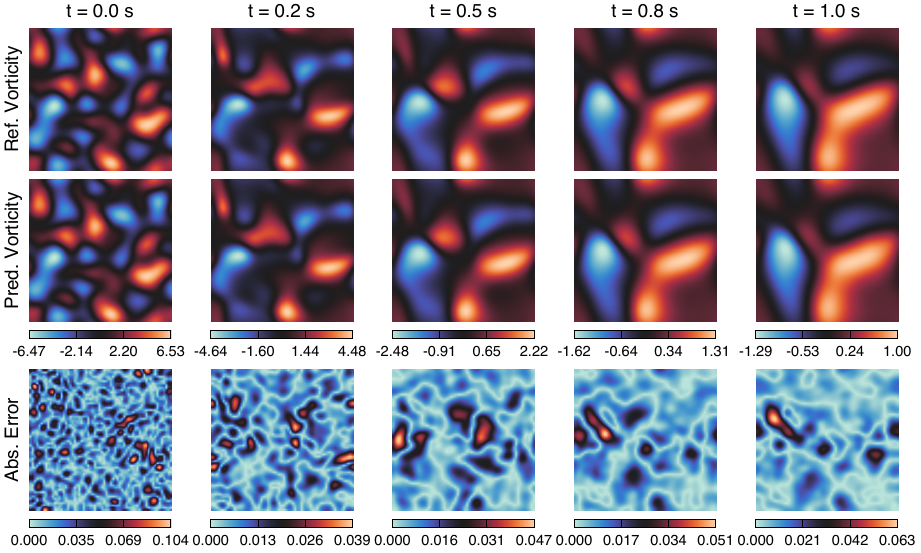}
	\caption{Burgers' equation. Vorticity field obtained via automatic differentiation from the velocity solution predicted by PI-CViT trained with a pure physics loss, compared with the reference vorticity computed via Fourier differentiation.}
	\label{fig:burgers_vorticity_field}
\end{figure}

\begin{figure}[h]
	\centering
	\includegraphics[width=\textwidth]{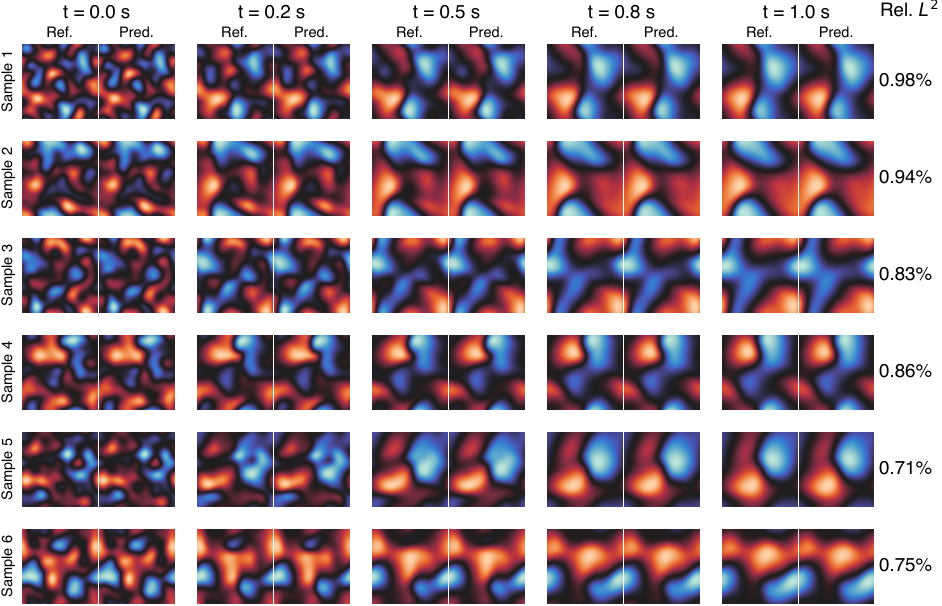}
	\caption{Burgers' equation. Velocity component $v_1$ predicted by PI-CViT trained with a pure physics loss for additional test samples with randomly sampled initial conditions.}
	\label{fig:burgers_solution_field_samples}
\end{figure}

\paragraph{Burgers' equation.}
We visualize the predicted and reference vorticity fields in Figure~\ref{fig:burgers_vorticity_field}. The solution fields with more randomly sampled initial conditions are shown in Figure~\ref{fig:burgers_solution_field_samples}.

\begin{figure}[h]
	\centering
	\includegraphics[width=\textwidth]{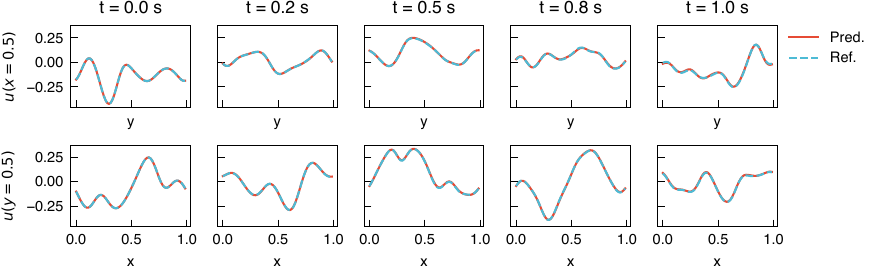}
\caption{{Wave equation.} Space slices ($x_1=0.5$ and $x_2=0.5$) of the scalar displacement $v$ predicted by PI-CViT trained with a pure physics loss and the reference solution at different time steps for a representative sample.}
	\label{fig:wave_solution_slices}
\end{figure}

\begin{figure}[h]
	\centering
	\includegraphics[width=\textwidth]{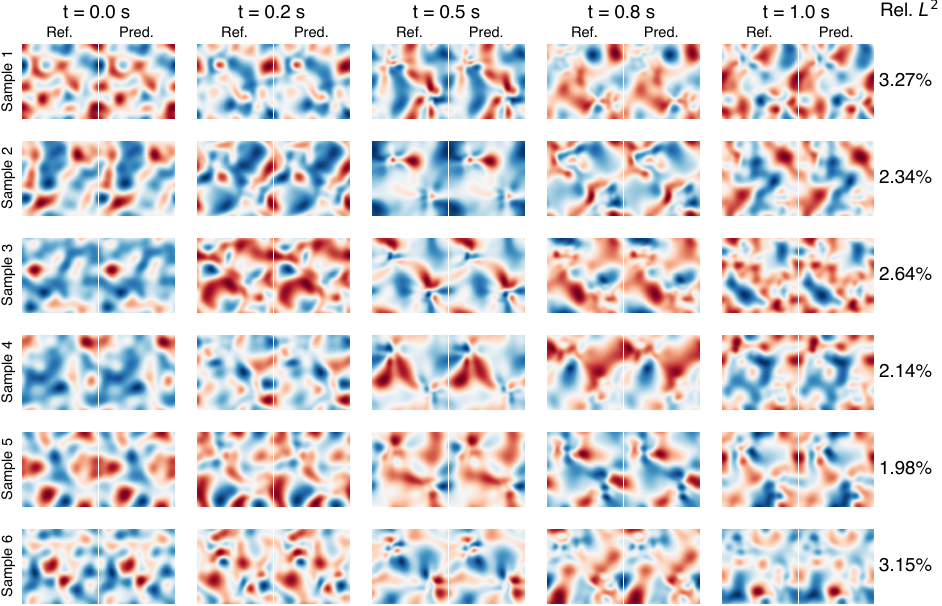}
	\caption{Wave equation. Scalar displacement $v$ predicted by PI-CViT trained with a pure physics loss for additional test samples with randomly sampled initial conditions.}
	\label{fig:wave_solution_field_samples}
\end{figure}

\paragraph{Wave equation.} We slice the predicted and reference solution fields along the vertical ($x_1=0.5$) and horizontal ($x_2=0.5$) centerlines at different time steps to visualize the spatiotemporal evolution of the solution in Figure~\ref{fig:wave_solution_slices}. The solution fields with more randomly sampled initial conditions are shown in Figure~\ref{fig:wave_solution_field_samples}.

\begin{figure}[h]
	\centering
	\subfigure[]{\includegraphics[width=\textwidth]{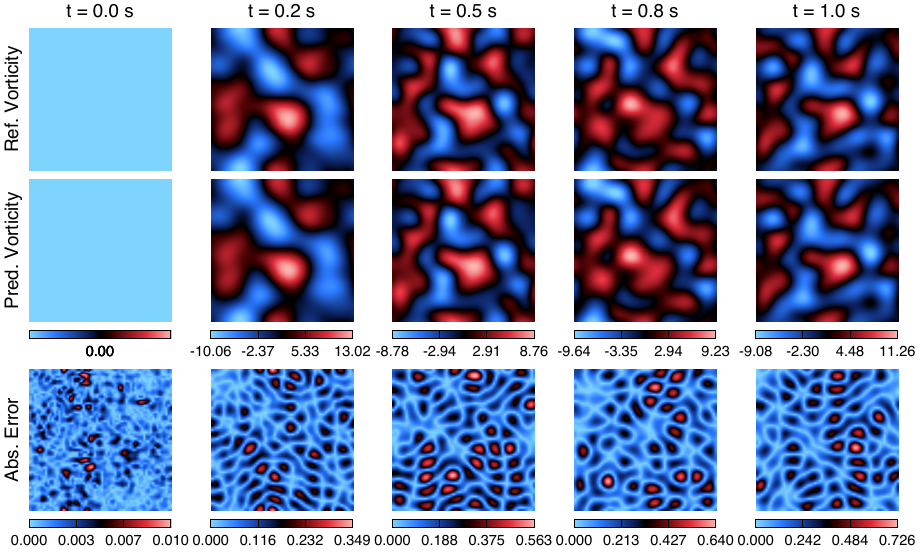}}\\
	\subfigure[]{\includegraphics[width=\textwidth]{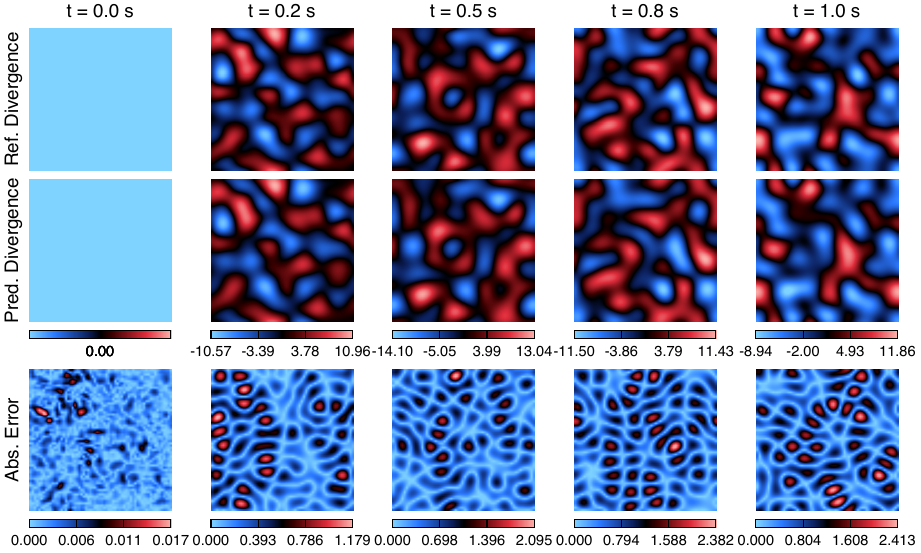}}
	\caption{Shallow water equations. (a)~Vorticity and (b)~divergence fields obtained via automatic differentiation from the velocity solution predicted by PI-CViT trained with a pure physics loss, compared with the reference fields computed via Fourier differentiation.}
	\label{fig:swe_vorticity_field}
\end{figure}

\begin{figure}[h]
	\centering
	\includegraphics[width=\textwidth]{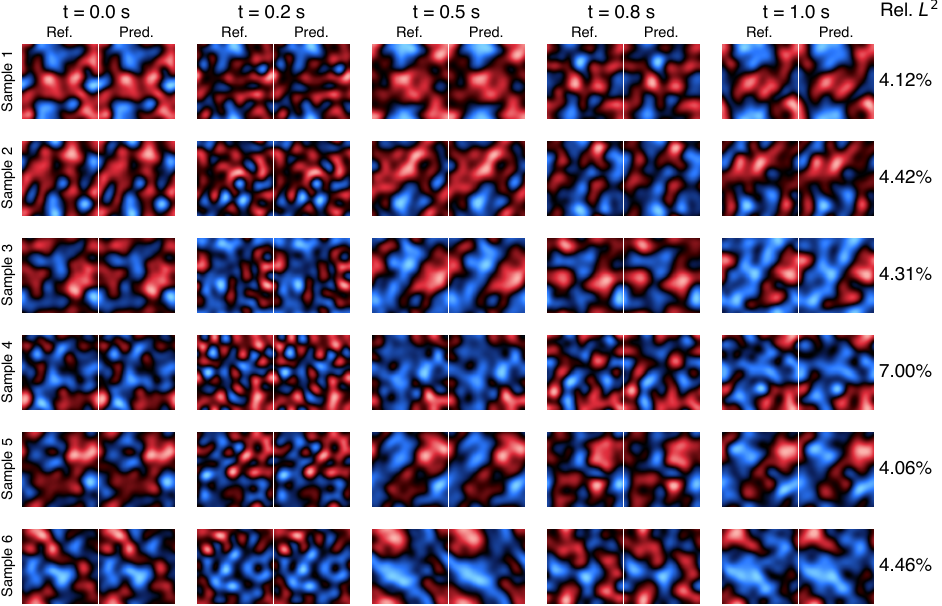}
	\caption{Shallow water equations. Free-surface height $h$ predicted by PI-CViT trained with a pure physics loss for additional test samples with randomly sampled initial conditions.}
	\label{fig:swe_solution_field_samples}
\end{figure}

\paragraph{Shallow water equations.} We visualize the predicted and reference vorticity and divergence fields in Figure~\ref{fig:swe_vorticity_field}, which emerge from the zero initial velocity field. The free-surface height $h$ for additional test samples with randomly sampled initial conditions is shown in Figure~\ref{fig:swe_solution_field_samples}.

\begin{figure}[h]
	\centering
	\includegraphics[width=\textwidth]{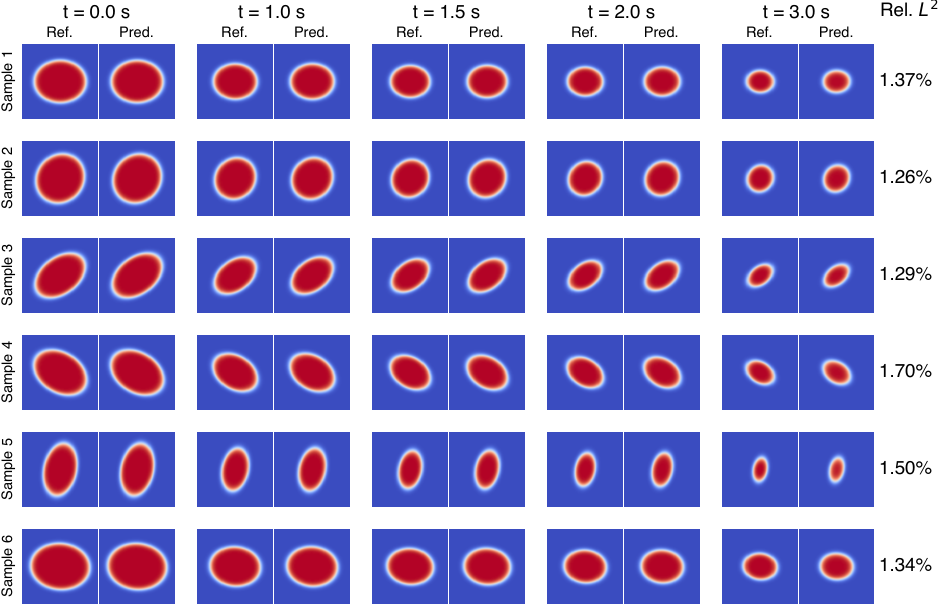}
	\caption{Ice melting problem. Phase-field variable $\phi$ predicted by PI-CViT trained with a pure physics loss for additional test samples with randomly sampled initial conditions.}
	\label{fig:ice_melting_solution_field_samples}
\end{figure}

\paragraph{Ice melting problem.} We provide the solution field with more randomly sampled initial conditions in Figure~\ref{fig:ice_melting_solution_field_samples}.


\begin{figure}[h]
	\centering
	\includegraphics[width=\textwidth]{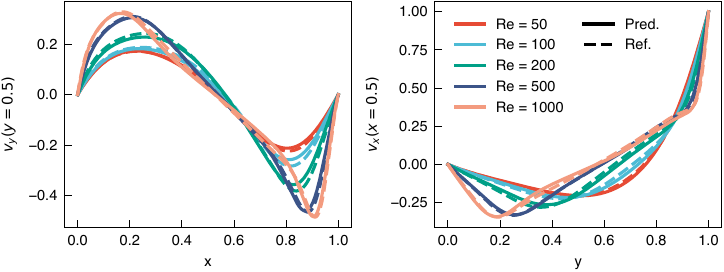}
	\caption{Lid-driven cavity flow. Velocity profiles along the vertical ($x_1=0.5$) and horizontal ($x_2=0.5$) centerlines predicted by PI-CViT trained with a pure physics loss, compared with the reference solution.}
	\label{fig:ldc_velocity_profile_all}
\end{figure}

\begin{figure}[h]
	\centering
	\includegraphics[width=\textwidth]{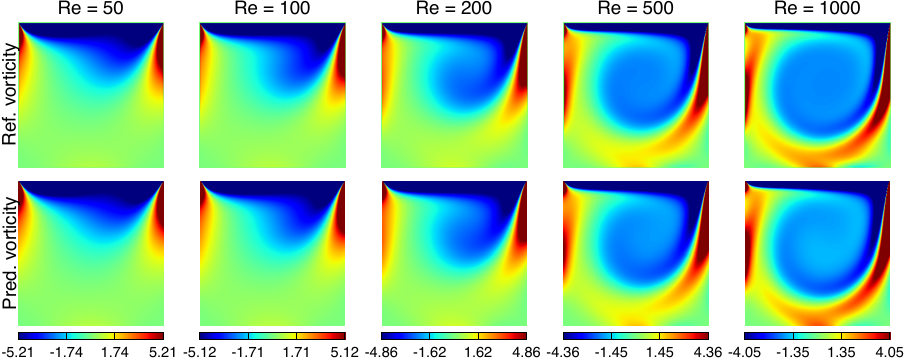}
	\caption{Lid-driven cavity flow. Vorticity field obtained via automatic differentiation from the velocity solution predicted by PI-CViT trained with a pure physics loss, compared with the reference vorticity.}
	\label{fig:ldc_vorticity}
\end{figure}

\paragraph{Lid-driven cavity flow.} We visualize the velocity $v_1$ along the vertical centerline ($x_1=0.5$) and velocity $v_2$ along the horizontal centerline ($x_2=0.5$) with varying Reynolds numbers in Figure~\ref{fig:ldc_velocity_profile_all}. The vorticity field is visualized in Figure~\ref{fig:ldc_vorticity}.

\section{Negative results}
\label{app:negative_results}

\begin{figure}[h]
	\centering
	\subfigure{
		\includegraphics[width=0.95\textwidth]{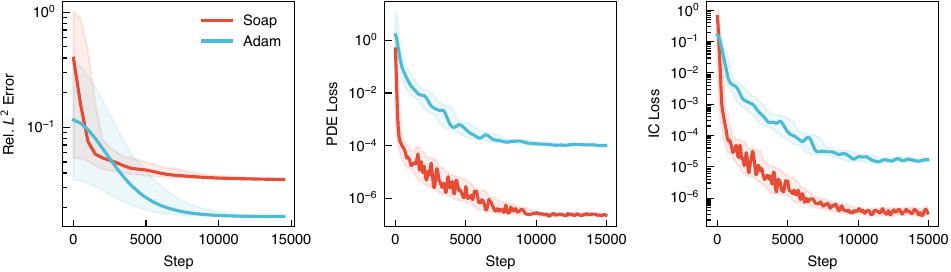}
	}


	\subfigure{
		\includegraphics[width=0.95\textwidth]{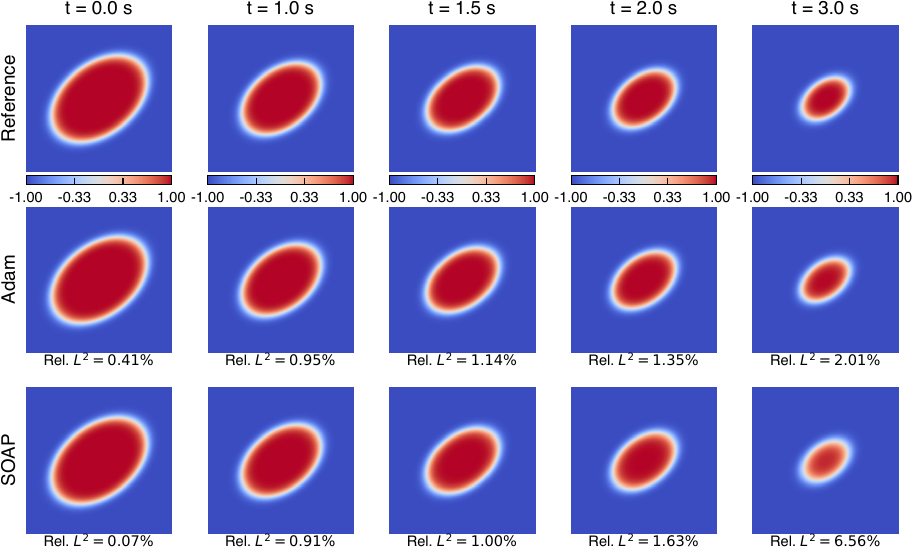}
	}
	\caption{Ice melting. Negative results from optimizer comparisons.
	\textbf{Top:} Training curves for relative $L^2$ error (left), PDE residual loss (middle), and initial condition loss (right). SOAP reduces both physics losses by roughly two orders of magnitude compared to Adam, yet converges to a higher relative $L^2$ error.
	\textbf{Bottom:} Predicted phase-field evolution. SOAP yields more accurate predictions at early time steps ($t \leq \qty{1.5}{s}$), but the interface progressively diffuses at later times, reaching \qty{6.56}{\percent} error at $t = \qty{3.0}{s}$ versus \qty{2.01}{\percent} for Adam. The discrepancy indicates that the physics loss preferentially rewards a diffuse interface, exposing a limitation of the current loss formulation for interface problems.}
	\label{fig:negative_optimizer}
\end{figure}

On the ice melting benchmark, SOAP produces a paradoxical outcome: despite driving both the PDE residual loss and the initial condition loss roughly two orders of magnitude below Adam, it yields a higher relative $L^2$ error (\qty{3.0}{\percent} vs.\ \qty{1.9}{\percent}).
As shown in Figure~\ref{fig:negative_optimizer}, SOAP achieves better pointwise accuracy at early time steps ($t \leq \qty{1.5}{s}$), but the predicted interface progressively diffuses at later times, losing the well-defined structure that characterizes the physical solution.

We attribute this to a structural mismatch between the loss and the evaluation metric.
A diffuse interface incurs smaller spatial gradients and thus a smaller PDE residual than a sharp but mislocated interface; the loss therefore inadvertently rewards diffusion as a way to reduce residuals.
Because SOAP is a more powerful optimizer, it converges more aggressively toward this diffuse-interface basin, achieving lower loss values at the cost of physical accuracy.
Adam, converging more slowly, settles at a sharper interface that better matches the reference.
The problem thus lies not with SOAP itself but with the loss formulation: the standard pointwise PDE residual does not penalize interface diffusion, so a blurrier prediction can appear better by the training objective while incurring higher $L^2$ error.
Potential remedies include more structured loss formulations, such as a variational form~\cite{eshaghi2025variational} or pseudo-time stepping~\cite{wang2026pinnsgowrong}, which we leave for future work.

\end{document}